\documentclass[12pt]{article}

\usepackage{scicite}

\usepackage{times}

\usepackage{graphicx}%
\usepackage{multirow}%
\usepackage{amsmath,amssymb,amsfonts}%
\usepackage{amsthm}%
\usepackage{mathrsfs}%
\usepackage[title]{appendix}%
\usepackage{xcolor}%
\usepackage{textcomp}%
\usepackage{manyfoot}%
\usepackage{booktabs}%
\usepackage{algorithm}%
\usepackage{algorithmicx}%
\usepackage{algpseudocode}%
\usepackage{listings}%

\usepackage{pdfpages}
\usepackage{caption}
\usepackage{epstopdf} 
\usepackage[backref]{hyperref}
\usepackage{caption}
\usepackage{stfloats}
\usepackage{float}
\usepackage{subfig}

\usepackage{adjustbox}
\usepackage{arydshln}

\usepackage{wrapfig}

\newenvironment{sciabstract}{%
\begin{quote} \bf}
{\end{quote}}

\title{A Novel Paradigm for Neural Computation: X-Net with Learnable Neurons and Adaptable Structure}

\author
{Yanjie Li,$^{13}$ Weijun Li,$^{123\ast}$ Lina Yu$^{1}$, Min Wu$^{1}$, Jingyi Liu$^{1}$, Wenqiang Li$^{13}$, \\ Linjun Sun$^{1}$, Meilan Hao$^{1}$, Shu Wei$^{13}$, Yusong Deng$^{13}$, LipingZhang$^{1}$, XiaoliDong$^{1}$,\\
HongQin$^{1}$, XinNing$^{1}$, YuguiZhang$^{1}$, BaoliLu$^{1}$, JianXu$^{1}$, ShuangLi$^{1}$
\\
\normalsize{$^{1}$AnnLab, Institute of Semiconductor, Chinese Academy of Sciences}\\
\normalsize{HaiDian, Beijing, 100083, CN}\\
\normalsize{$^{2}$School of Integrated Circuits, University of Chinese Academy of Sciences}\\
\normalsize{HuaiRou, Beijing, 101408, CN}\\
\normalsize{$^{3}$School of Electronic, Electrical and Communication Engineering }\\
\normalsize{University of Chinese Academy of Sciences }\\
\normalsize{HuaiRou, Beijing, 101408, CN}\\
\\
\normalsize{$^\ast$To whom correspondence should be addressed; E-mail:  wjli@semi.ac.cn.}
}

\date{}

\begin{document} 


\baselineskip24pt


\maketitle


\begin{sciabstract}
Multilayer perception (MLP) has permeated various disciplinary domains, ranging from bioinformatics to financial analytics, where their application has become an indispensable facet of contemporary scientific research endeavors. However, MLP has obvious drawbacks. 1), The type of activation function is single and relatively fixed, which leads to poor `representation ability' of the network, and it is often to solve simple problems with complex networks; 2), the network structure is not adaptive, it is easy to cause network structure redundant or insufficient. In this work, we propose a novel neural network paradigm X-Net promising to replace MLPs. X-Net can dynamically learn activation functions individually based on derivative information during training to improve the network's representational ability for specific tasks. At the same time, X-Net can precisely adjust the network structure at the neuron level to accommodate tasks of varying complexity and reduce computational costs. We show that X-Net outperforms MLPs in terms of representational capability. X-Net can achieve comparable or even better performance than MLP with much smaller parameters on regression and classification tasks. Specifically, in terms of the number of parameters, X-Net is only 3$\%$ of MLP on average, and only 1.1$\%$ under some tasks. 
We also demonstrate X-Net's ability to perform scientific discovery on data from various disciplines such as energy, environment, and aerospace, where X-Net is shown to help scientists discover new laws of mathematics or physics.
\end{sciabstract}

\section*{Introduction}
Multi-Layer Perceptron (MLP) is the cornerstone of contemporary artificial intelligence. In the field of artificial intelligence, the study of MLP\textsuperscript{\cite{ann1,ann2,ann3,ann4,ann5}} is nearly as old as AI itself. However, there is an irreconcilable contradiction between the scale and performance of the MLP. To obtain better performance, it is often necessary to continuously expand the scale of the network in terms of depth and breadth. For example, the performance of GPT-1 to GPT-4\textsuperscript{\cite{gpt1,gpt2,gpt3,gpt3.5,gpt4}} is getting better and better, but the number of parameters rapidly increases from 1.3B to a trillion level, which brings the problem of energy consumption, calculation, storage, communication, and other costs\textsuperscript{\cite{de2023growing}}. The high cost will hinder its application and promotion. 
So why do current neural networks tend to be so big? Its technical roots lie in two points:

1), The neuron activation function of classical neural networks is single and fixed, and its representation ability is relatively insufficient. Many neurons are often required to fit other types of nonlinear functions. Moreover, as the dimensionality of the problem increases, the number of neurons explodes exponentially.

2), The network structure as a hyperparameter needs to be predetermined and kept fixed during training. But in fact, it is difficult for us to rely on artificial experience to obtain an optimal network structure, and it is easy to have structural redundancy or insufficiency.
\begin{figure}[t]
\vspace{0.4cm}
\begin{center}
\centerline{\includegraphics[width=1.0\linewidth]{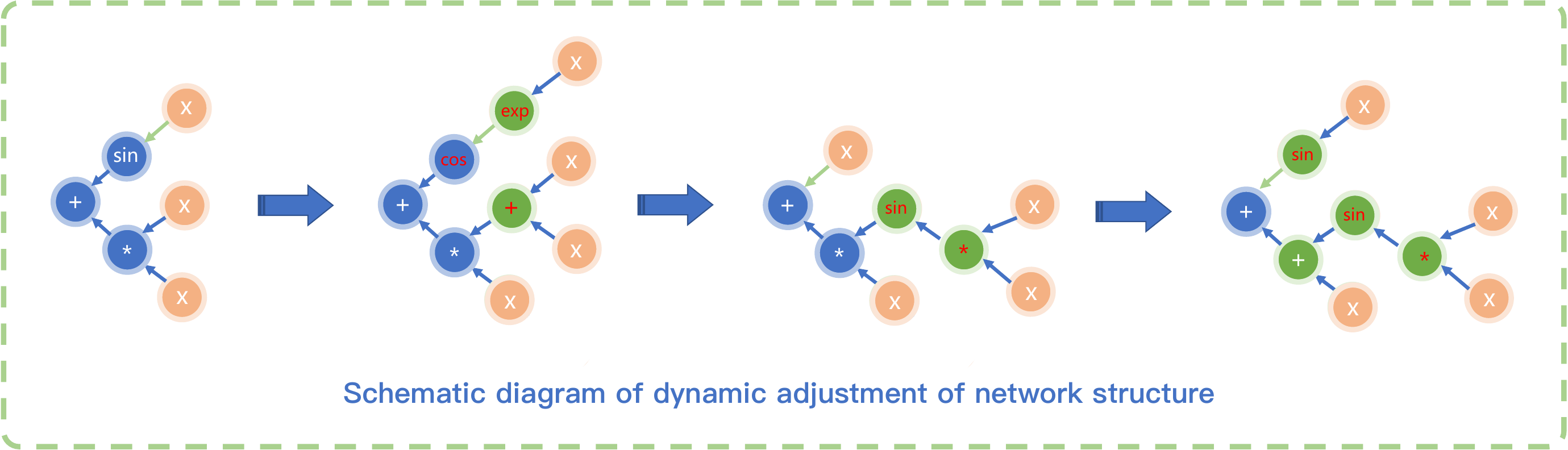}}
\caption{Figure illustrates the dynamic changes in the network structure of X-Net during the training process.}
\label{fig11}
\end{center}
\end{figure}

\textbf{So, it is necessary to study a new generation of neural networks with dynamic learnable activation functions and adaptive adjustment of network structure}. This is also a problem that future artificial intelligence research must face. Because human computing power is ultimately limited, the scale of the model cannot be expanded indefinitely.

In order to overcome the above problems, we try to explore a new generation of neural networks. We propose a new neural network called X-Net. 
\begin{table}[!h]
\centering
\caption{The characteristics of X-Net and MLPs.}
\label{tab:char}
\begin{tabular}{lcccc}
\toprule
& Neuron Type & Neuron Learnability & Weight Learnability & Structure\\
\midrule
MLPs & Single & \rlap{\texttimes} & \rlap{\checkmark} & Pre-defined\\
X-Net & Multiple & \rlap{\checkmark} & \rlap{\checkmark} & Self-learnable\\
\bottomrule
\end{tabular}
\end{table}
Table\ref{tab:char} shows the characteristics of X-Net and MLP. The X-Net theoretically can use any differentiable function as the activation function, The experimental results show that the representation ability is greatly improved compared with MLP\ref{Representation}. The type of activation function can be learned in real-time according to the needs of the task under the guidance of the gradient during the network training.
In particular, The network structure is dynamically adjusted in real-time, which makes the network structure match the task requirements accurately and reduces the occurrence of redundancy and insufficiency. Fig. \ref{fig11} shows a schematic diagram of the training process of X-Net. 

In addition, we found that when the activation function uses some mathematical notation with nice properties (e.g., sin() has periodicity, etc.), X-Net can find some interpretable mathematical formulas in some cases. This will greatly improve the problem of uninterpretable results of traditional neural networks. It also shows that X-Net has great potential in facilitating scientific discovery. We believe this will be extremely attractive to researchers from other disciplines (biology, physics, materials, etc.) and we think it will drive the field of AI for Science to flourish even more.

In summary, the X-Net model not only opens up a new research direction in the field of neural networks but also sets up a new technical standard for building truly adaptive intelligent systems. In addition, X-Net has good universality and can empower the development of various disciplines. It has a very broad research prospect and research space. Finally, we hope that our study will provoke a reflection on the inadequacy of current neural network architectures, as well as generate enthusiasm for the exploration of next-generation neural networks.
\section*{Results}

In order to comprehensively evaluate the performance of X-Net, we make a comprehensive evaluation of X-Net on regression and classification tasks. In particular, we also test X-Net's ability to make scientific discoveries on datasets from various disciplines such as environment and energy.

\subsection{Representation capability with learnable activation functions}\label{Representation}
In order to compare the representation capabilities of X-Net and MLPs, we designed a unique experiment: we initialized the X-Net and MLP with three layers of neurons with 4, 2, and 1 neurons per layer. The difference is that the different layers in X-Net are sparsely connected in the form of a binary tree, whereas the MLP is fully connected. Each neuron in X-Net has a different activation function, whereas neurons in MLP all have a single ReLU activation function. We optimize X-Net and MLP using the LBFGS. The results in Table~\ref{tab:repre} show that X-Net exhibits a better fitting ability on the Nguyen dataset compared to MLPs, despite being sparsely connected. This result demonstrates that the X-Net can achieve superior nonlinear representation ability than the fully connected MLPs due to the diversity of activation functions. This also proves that the hypothesis of improving the representation ability of neural networks by increasing the type of activation function is feasible.

\begin{table}[!h]
\centering
\caption{The representation ability of using multiple activation function types versus a single activation function type for the same number of neurons.
\label{tab:repre}}
\begin{tabular}{cccc}
\toprule
Data& Benchmark&\multicolumn{2}{c}{Activation Type}\\ 
\toprule
& &Multiple &Single  \\ 
\cmidrule(lr){1-2}
\cmidrule(lr){3-4}

Nguyen-1 & $x_1^3+x_1^2+x_1 $ & 0.9999&0.5924 \\
Nguyen-2 & $x_1^4+x_1^3+x_1^2+x_1$& 0.9999&0.3922  \\
Nguyen-3 & $x_1^5+x_1^4+x_1^3+x_1^2+x_1$& 0.9999&0.7645  \\
Nguyen-4 & $x_1^6+x_1^5+x_1^4+x_1^3+x_1^2+x_1$& 0.9995&0.8023 \\
Nguyen-5 & $\sin(x_1^2)\cos(x)-1$&0.9956&0.2217  \\
Nguyen-6 & $\sin(x_1)+\sin(x_1+x_1^2)$& 0.9995& 0.4337 \\
Nguyen-7 & $\log(x_1+1)+\log(x_1^2+1)$& 0.9999 &0.6902 \\
Nguyen-8 & $\sqrt{x}$& 0.9999& 0.6756 \\
Nguyen-9 & $\sin(x)+\sin(x_2^2)$  & 0.9940 &0.7726 \\
Nguyen-10& $2\sin(x)\cos(x_2)$& 0.9859&0.8126  \\
Nguyen-11 & $x_1^{x_2}$ & 0.9879&0.6673 \\
Nguyen-12 & $x_1^4-x_1^3+\frac{1}{2}x_2^2-x_2$& 0.9824&0.8615 \\ 
\cmidrule(lr){1-2}
\cmidrule(lr){3-4}
&Average &\textbf{0.9954} & \textbf{0.6572 }\\
\toprule
\end{tabular}
\end{table}

\subsection{Complexity compared with MLPs}
Artificial neural networks have a fixed network structure and a number of nodes, which can easily lead to redundancy in parameters and nodes, greatly slowing down the convergence efficiency of the neural network. In contrast, X-Net is more flexible, with a dynamically changing network structure and the number of nodes that can be adaptively adjusted based on the complexity of the problem. This greatly alleviates the problem of redundancy in nodes and parameters in ordinary neural networks. To test the complexity of the model obtained by four algorithms X-Net, and MLP on the same task, we stop the training when $R^2=0.99$, and count the number of nodes and parameters used by the two networks at this time. The specific statistical results are shown in Table \ref{tab1}.
From Table \ref{tab1}(REGRESSION), it is evident that in fitting the Nguyen data set, the MLP necessitates approximately fourfold the node count in comparison to X-Net and demands almost a twentyfold increase in the number of parameters for optimization. 
The results in Table \ref{tab1}(CLASSION) show that X-Net requires an average number of nodes comparable to that of MLP for the classification task, but the number of weight parameters is only about 1.4\% of that of MLP, and under these conditions, X-Net exhibits performance comparable to that of MLP.

\subsection{Performance on regression tasks}
\subsubsection{Fitting accuracy}

\begin{table*}[ht]
\center
\caption{Comparison of various indicators between X-Net and MLP on regression and Classion tasks
\label{tab1}}
\resizebox{\textwidth}{64mm}{
\begin{tabular}{cccccccc}
\multicolumn{8}{c}{REGRESSION}\\
\toprule[1.45pt]
\toprule[1pt]
Data&Benchmark& \multicolumn{3}{c}{X-Net}& \multicolumn{3}{c}{MLP}\\  
\cmidrule[0.1pt]{1-8}
& &Fitting($R^2$)  &Nodes & Parameters  & Fitting($R^2$) &Nodes & Parameters \\ 
\cmidrule(lr){1-2}
\cmidrule(lr){3-5}
\cmidrule(lr){6-8}
Nguyen-1 & $x_1^3+x_1^2+x_1 $  &  $1.0000_{\pm0.00} $   &5 &12 &$0.9999_{\pm0.08}$  &14 &78 \\
Nguyen-2 & $x_1^4+x_1^3+x_1^2+x_1$& $ 1.0000_{\pm0.00}$  &9&38  & $0.9999_{\pm0.04} $&18 &118\\
Nguyen-3 & $x_1^5+x_1^4+x_1^3+x_1^2+x_1$  & $ 0.9999_{\pm0.06}$  &14 &58&$ 0.9999_{\pm0.05} $&18 &118\\
Nguyen-4 & $x_1^6+x_1^5+x_1^4+x_1^3+x_1^2+x_1$&$  0.9999_{\pm0.09} $ &20 &82 &$ 0.9998_{\pm0.10} $&28 &253\\
Nguyen-5 & $\sin(x_1^2)\cos(x)-1$&$  0.9998_{\pm0.04} $ &5 &16 &$0.9984_{\pm0.08}$  &26 &222 \\
Nguyen-6 & $\sin(x_1)+\sin(x_1+x_1^2)$& $  1.0000_{\pm0.00}$  &6 &18&$ 0.9996_{\pm0.07}$ &8 &333  \\
Nguyen-7 & $\log(x_1+1)+\log(x_1^2+1)$ &$  0.9996_{\pm0.02} $ &5 &16 &$ 0.9998_{\pm0.06}$  &12 &61 \\
Nguyen-8 & $\sqrt{x}$ &$  1.0000_{\pm0.00}$  &1 &4 &$ 0.9999_{\pm0.01}$  &16 &97\\
Nguyen-9 & $\sin(x)+\sin(x_2^2)$&$  1.0000_{\pm0.00}$   &4 &14 & $0.9984_{\pm0.05}$ &40 &481\\
Nguyen-10 & $2\sin(x)\cos(x_2)$  & $1.0000_{\pm0.00}$  &7 &24& $0.9994_{\pm0.09}$   &130 &4468\\
Nguyen-11 & $x_1^{x_2}$& $1.0000_{\pm0.00} $& 3 & 10 & $0.9999_{\pm0.02}$ &16 &97\\
Nguyen-12 & $x_1^4-x_1^3+\frac{1}{2}x_2^2-x_2$& $0.9996_{\pm0.09}$ &9 &40 &$ 0.9988_{\pm0.20}$ &16 &97\\ 
\cmidrule(lr){1-2}
\cmidrule(lr){3-5}
\cmidrule(lr){6-8}
&Average & 0.9999 &7.33 &28.50& 0.9995 &28.50 &510.25\\
\bottomrule
\toprule
\multicolumn{8}{c}{CLASSION}\\
\toprule
Data&Benchmark& \multicolumn{3}{c}{X-Net}& \multicolumn{3}{c}{MLP}\\ 

\toprule
& &Accuracy& Nodes & Parameters & Accuracy& Nodes& Parameters  \\ 
\cmidrule(lr){1-2}
\cmidrule(lr){3-5}
\cmidrule(lr){6-8}

Dataset-1 &Iris &98.7\% & 28 & 112 & 99.0\%&67 & 1315  \\
Dataset-1 &Mnist(6-dim) &89.4\%& 65 & 244 & 88.6\%& 298 & 23342 \\
Dataset-1 &Mnist &99.5\%& 816 & 3084 & 99.2\%& 788 & 267612 \\ 
Dataset-1 &Fashion-MNIST(6-dim)&76.2\%& 122 & 486 & 75.1\%& 398 & 31143 \\
Dataset-1 &Fashion-MNIST&94.1\%&  1066& 3884 & 94.4\%& 1244 & 544129 \\
Dataset-1 &CIFAR-10(6-dim) &26.4\%& 206 & 764 & 24.6\%& 414 & 35292 \\
Dataset-1 &CIFAR-10 &46.8\%& 2733 & 10072 & 48.4\%& 2164 & 876503 \\ 
\cmidrule(lr){1-2}
\cmidrule(lr){3-5}
\cmidrule(lr){6-8}
&Average & 75.87\% & 719.43 & 2663.71& 75.61\% &767.57& 234190.86\\
\bottomrule
\end{tabular}
}
\end{table*}

We tested the above four algorithms on the Nguyen data set. The dataset contains 12 curves, each of which is a mathematical formula. See the appendix for details.
We used $R^2$ to test the fitting ability of our algorithm and MLP. 
The formula for $R^2$ is as follows\eqref{r2}.
\begin{equation}
\label{r2}
\displaystyle R^2 = 1-\frac{\sum_{i=0}^{N}{(y_i-\hat{y}_i)^2}}{\sum_{i=0}^{N}{(y_i-\overline{y})^2}}.
\end{equation}
where $y_i$ is the true label value of the $i^{th} $sampling point,  $\hat{y}_i$ is the value predicted by the model for the $i^{th} $ data, $\overline{y}$ is the mean of the true values $y$. 
The closer $R^2$ is to 1 the better the algorithm fits the target curve, and conversely, the farther $R^2$ is from 1, the worse the algorithm fits the target curve. The specific results are shown in Table \ref{tab1} 
(REGRESSION). From the table, we can clearly find that our algorithm has better inherited the powerful nonlinear fitting ability of the neural network, and its fitting ability is not weaker than that of excellent regression algorithms such as MLP.


\subsection{Performance on classification tasks}
The classification task stands as a quintessential sub-domain within the realm of machine learning.  Consequently, gauging the classification performance of algorithms becomes imperative.  To validate the classification efficacy of X-Net, we employed the Iris dataset, MNIST, Fishion-MNIST, and CIFAR-10 dataset as our experimental datasets.  We juxtaposed the performance of X-Net with the MLP. On the MNIST, Fishion-MNIST, and CIFAR-10 datasets, we conducted two distinct experiments: initially, we employed PCA to dimensionally reduce the data, transmuting the images from a p dimension(Number of pixels) down to 6 dimensions, and subsequently classifying using the reduced data. The secondary experiment involved direct classification without any dimension reduction. 

Table \ref{tab1} (CLAASSION) meticulously delineates the performance comparison between X-Net and MLP on the Iris, MNIST, Fashion-MNIST, and CIFAR-10 datasets. The outcomes suggest that X-Net marginally outperforms the conventional MLP across all seven classification tasks.
Regarding network structural complexity, X-Net surpasses its counterpart. Taking the Iris dataset as a case in point, the neuron count of X-Net is merely half that of MLP, with its parameter magnitude being a mere tenth of that of MLP.  On the MNIST dataset, In the experimentation employing PCA, the node count of X-Net is but a quarter of MLP, and its parameter magnitude stands at a mere $1.04\%$ (1/96) of MLP. 
In experiments without dimension reduction, the node count of X-Net amounts to two-thirds that of MLP, while the parameter count is a mere $1.1 \% $ (1/90) of MLP. The same trend is also shown in Fishion-MNIST and CIFAR-10. 

All in all, X-Net achieves comparable accuracy to MLP on classification tasks; however, its network structure complexity is much lower than MLP.
\subsection{Used in scientific discovery}
\subsubsection{Modeling in space science (Airfoil-self-noise)\textsuperscript{\cite{airfoil}}}
We use the airfoil-self-noise data set to conduct practical tests on the algorithms for our paper. The airfoil-self-noise data set consists of six dimensions, of which five dimensions are feature data, namely Frequency ($x_1$), Angle of attack ($x_2$), Chord length ($x_3$), Free-stream velocity ($x_4$), and Suction side displacement thickness ($x_5$). Our goal is to predict the Scaled sound pressure level ($Y$) data through the above five features. The formula discovered by applying our algorithm is shown in the following text, and the two equations we found are presented below (rounded to two decimal places):
\begin{figure*} 
    \centering
  
    \subfloat[Results of fitting the Airfoil-self-noise data (Eq.5)]{
        \includegraphics[width=0.3\linewidth]{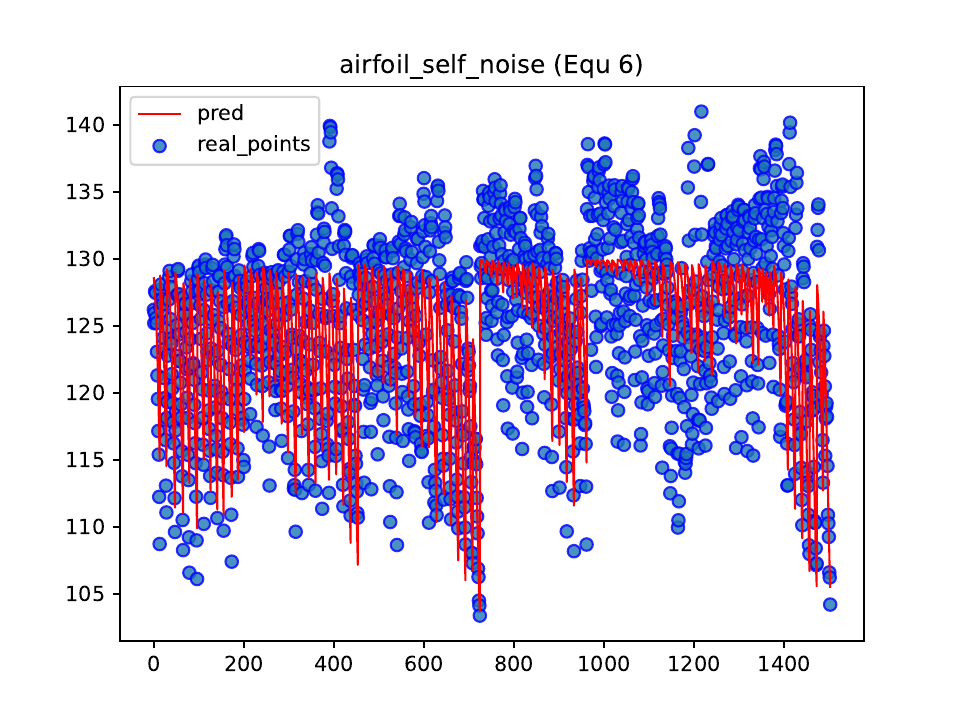}
         \label{2a}}
    \subfloat[Figure a after sorting by $y_{pre}$]{
        \includegraphics[width=0.3\linewidth]{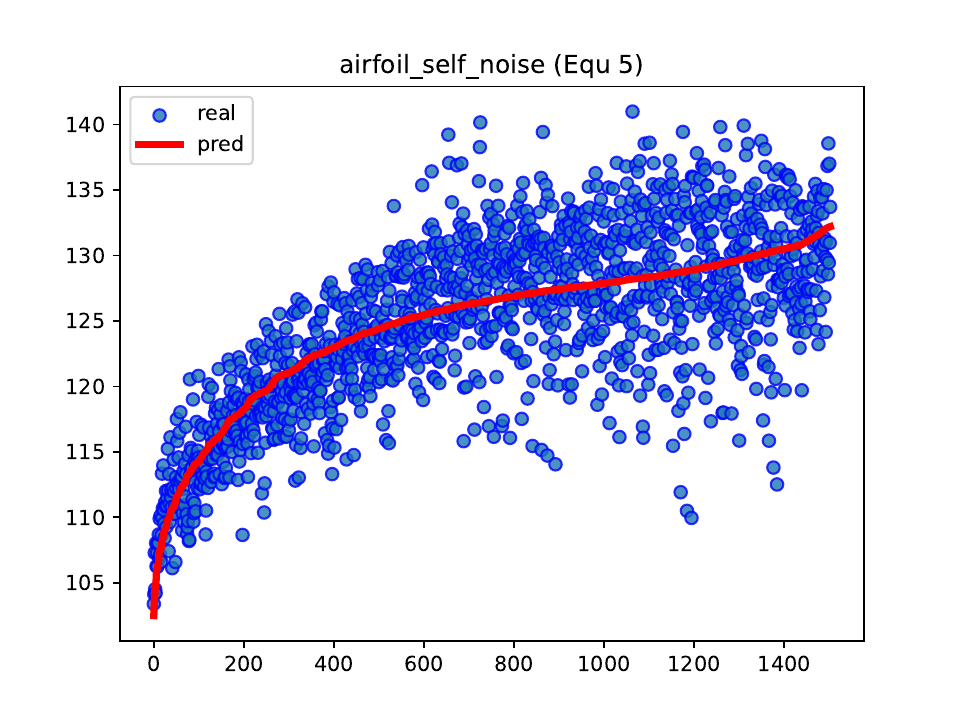}
    \label{2b}}
    \subfloat[Correlation coefficient matrix for Airfoil-self-noise]{
        \includegraphics[width=0.30\linewidth]{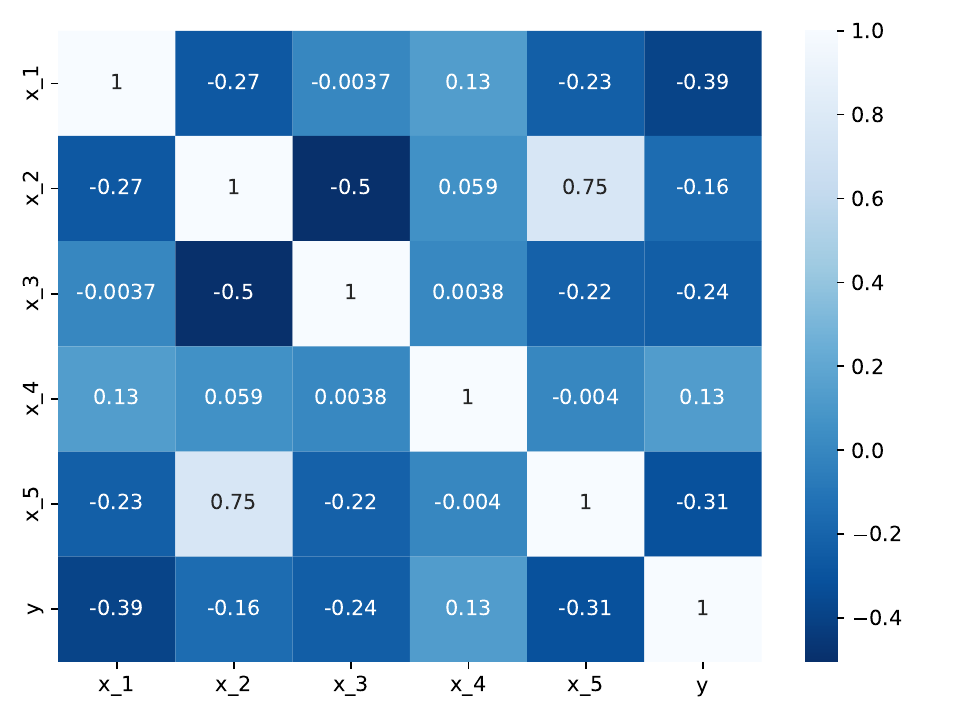}
    \label{2c}}
    \\
    \subfloat[Results of fitting the Airfoil-self-noise data(Eq.6)]{
        \includegraphics[width=0.3\linewidth]{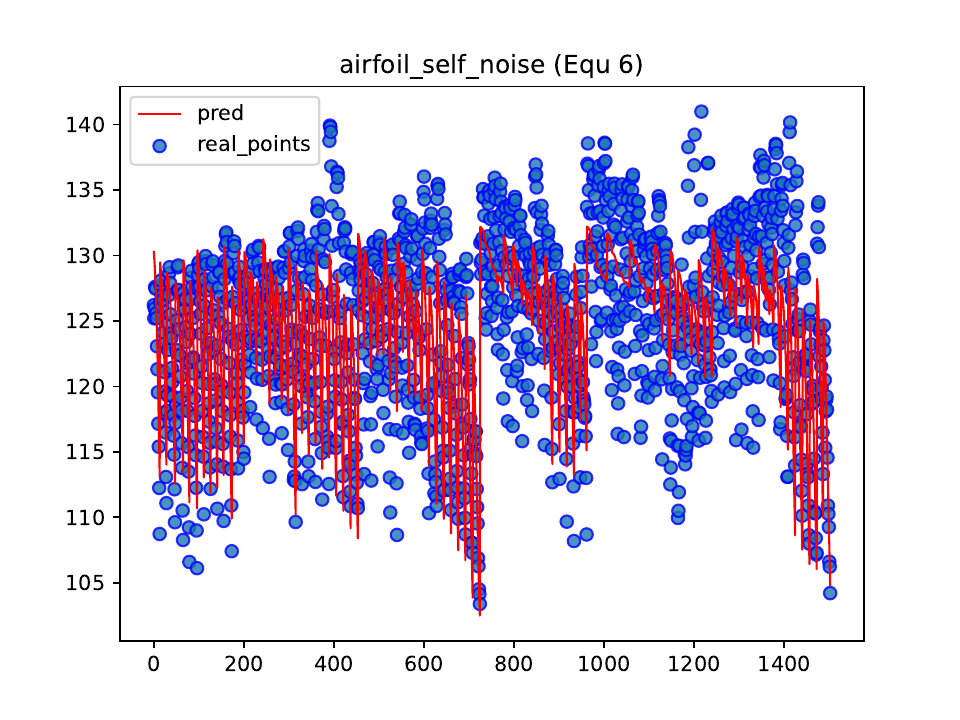}
    \label{2d}}
    \subfloat[Figure d after sorting by $y_{pre}$]{
        \includegraphics[width=0.3\linewidth]{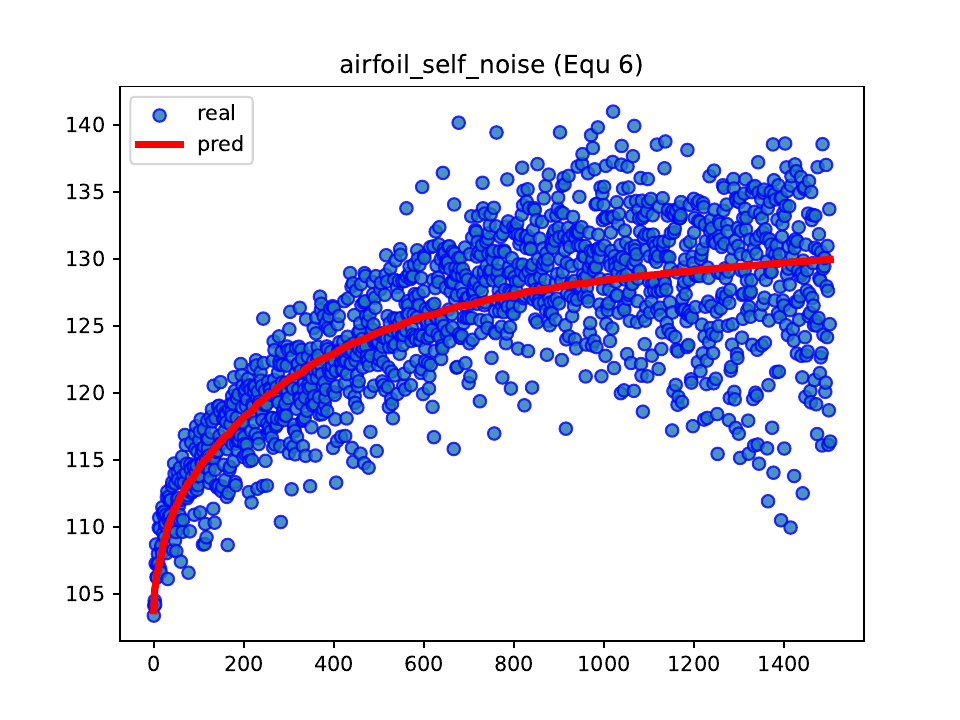}
    \label{2e}}
    \subfloat[Global temperature change prediction outcomes]{
       \includegraphics[width=0.3\linewidth]{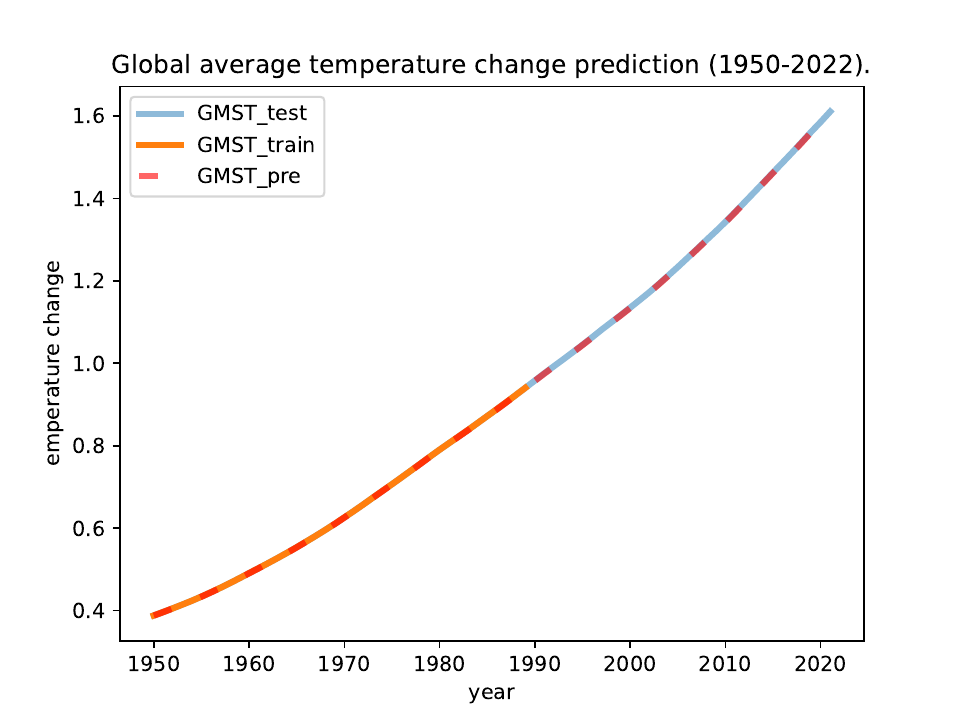}
    \label{2f}}\\
    \subfloat[Nguyen-1]{
      \includegraphics[width=0.11\linewidth]{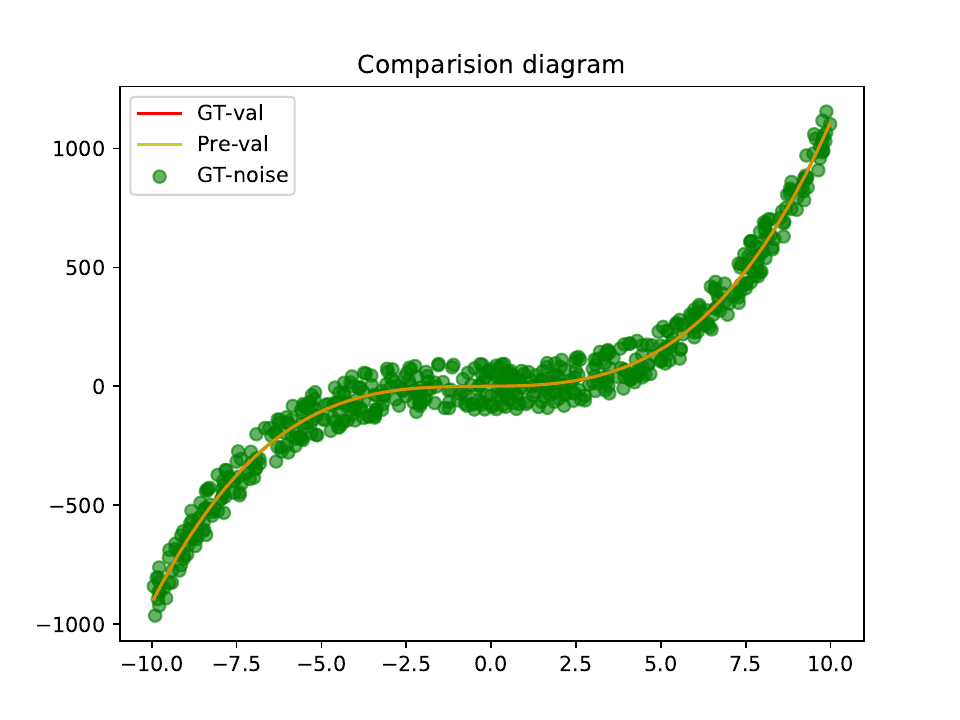}
    \label{1g}}
	  \subfloat[Nguyen-2]{
        \includegraphics[width=0.11\linewidth]{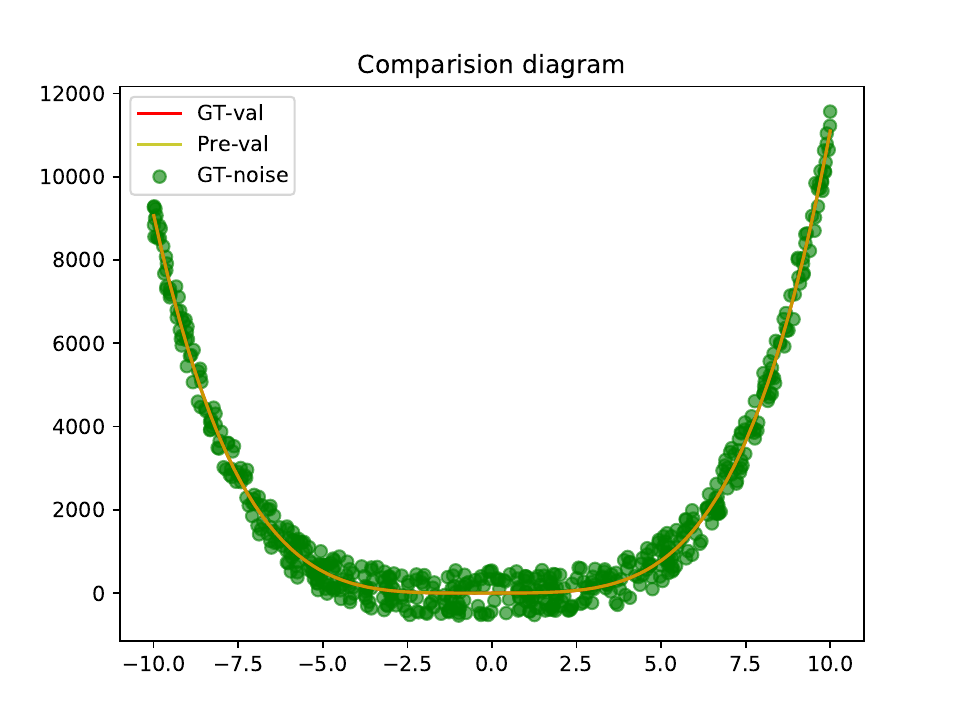}
    \label{1h}}
    \subfloat[Nguyen-3]{
        \includegraphics[width=0.11\linewidth]{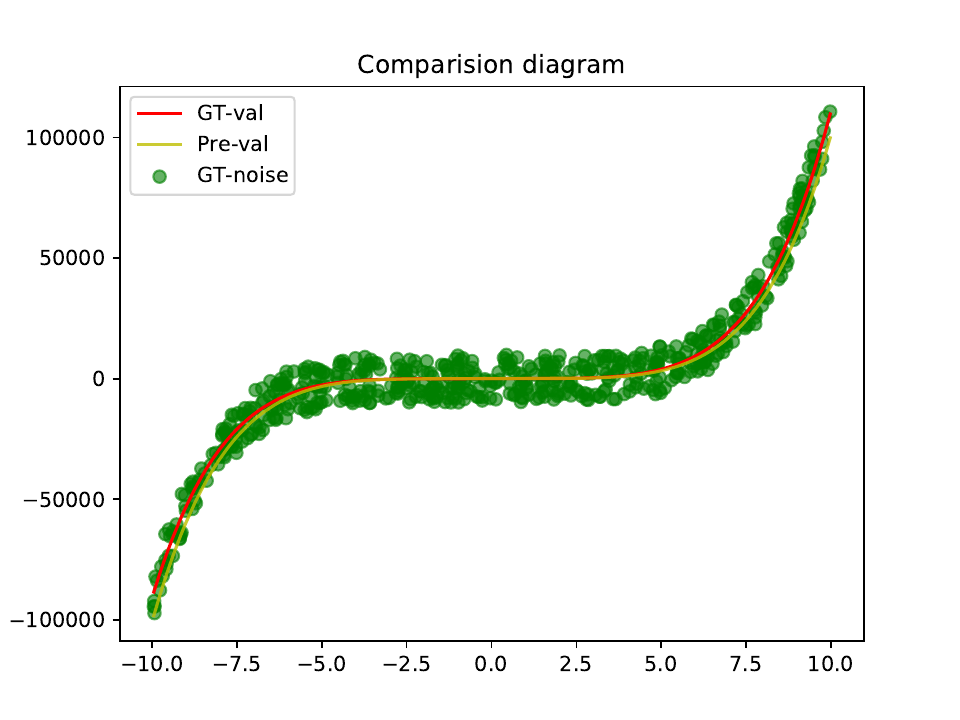}
    \label{1i}}
    \subfloat[Nguyen-4]{
        \includegraphics[width=0.11\linewidth]{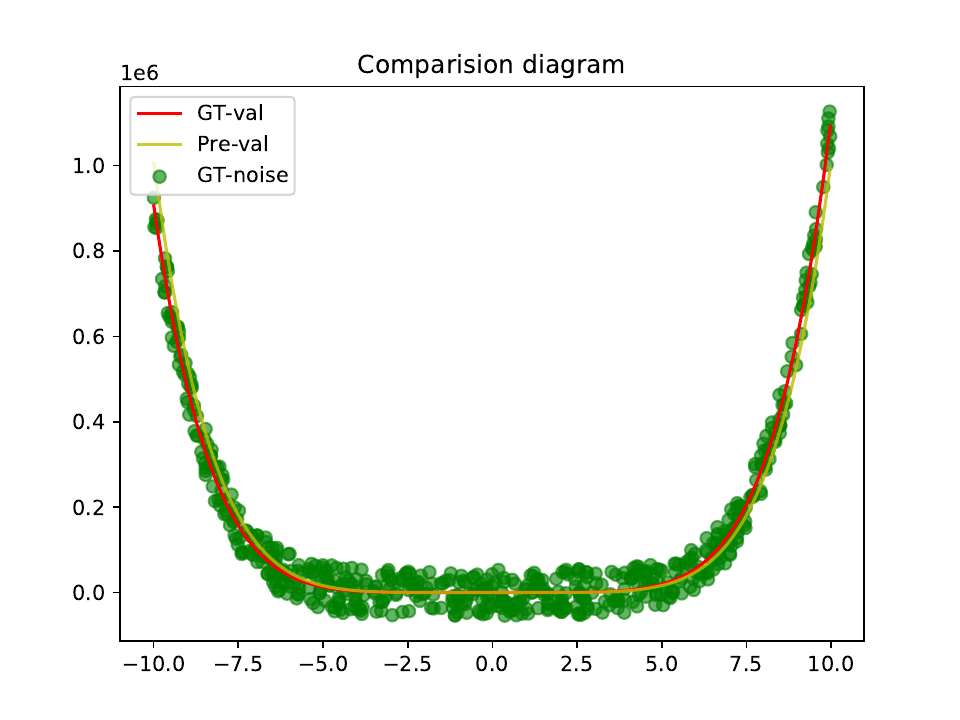}
    \label{1j}}
    \subfloat[Nguyen-5]{
      \includegraphics[width=0.11\linewidth]{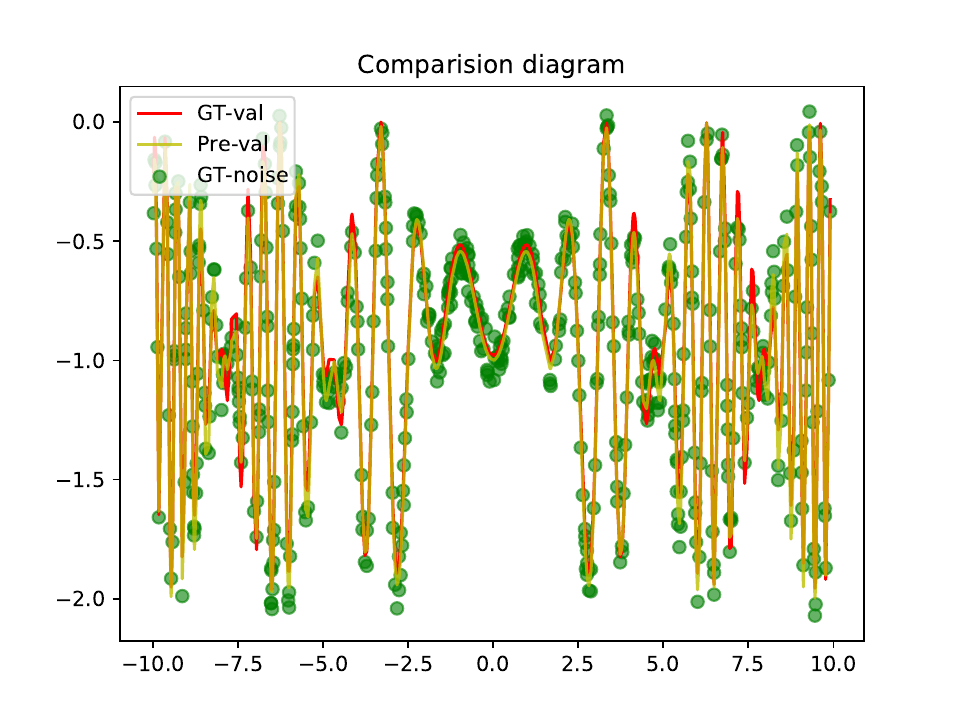}
    \label{1k}}
	  \subfloat[Nguyen-6]{
        \includegraphics[width=0.11\linewidth]{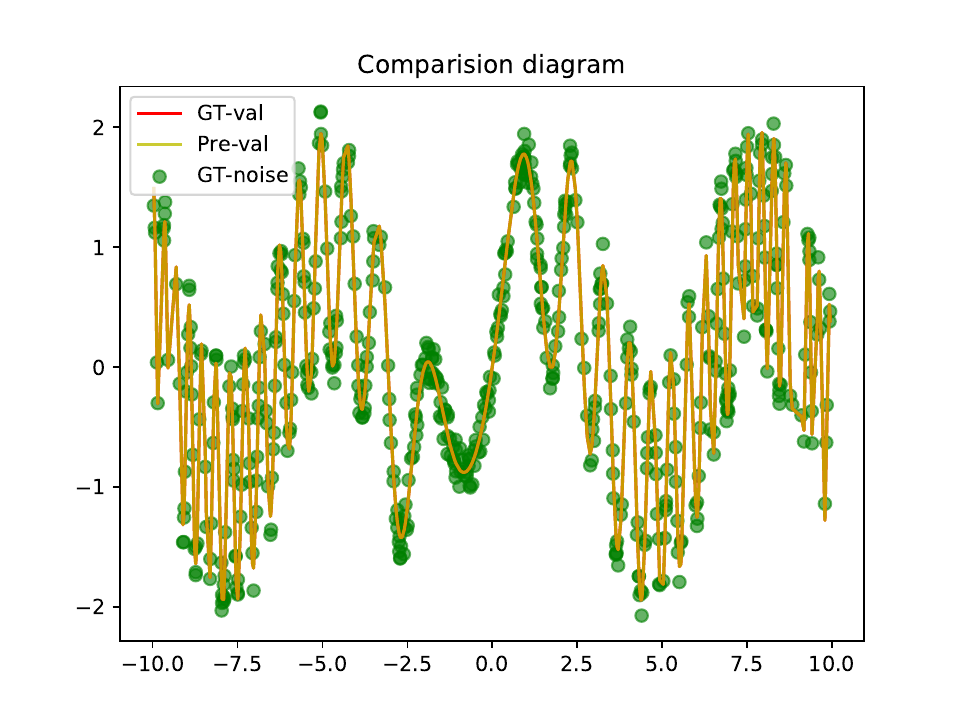}
    \label{1l}}
    \subfloat[Nguyen-7]{
        \includegraphics[width=0.11\linewidth]{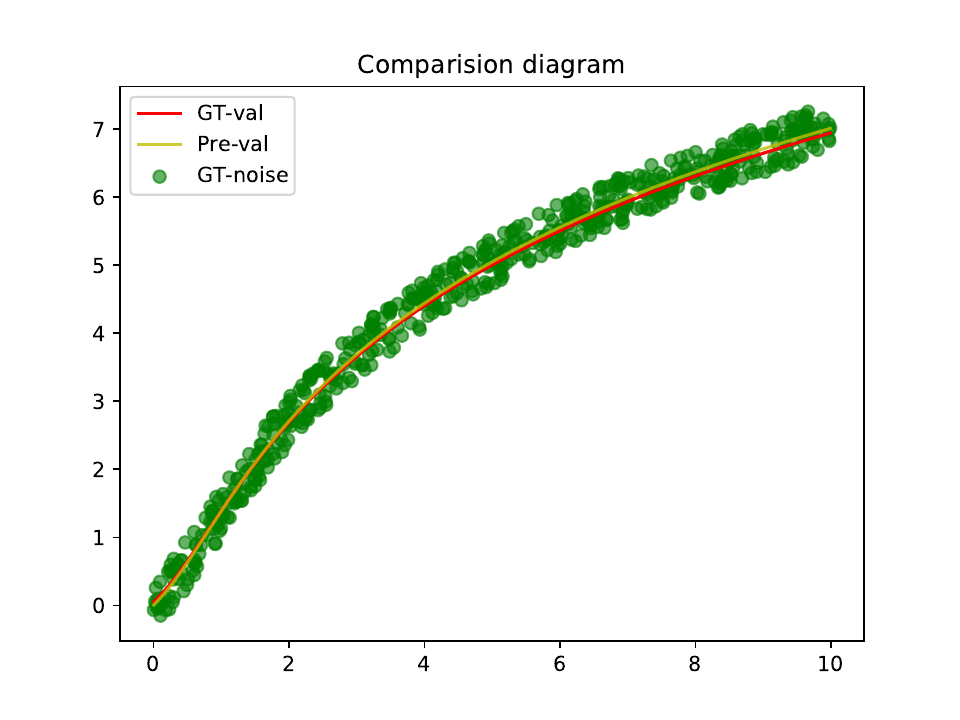}
    \label{1m}}
    \subfloat[Nguyen-8]{
        \includegraphics[width=0.11\linewidth]{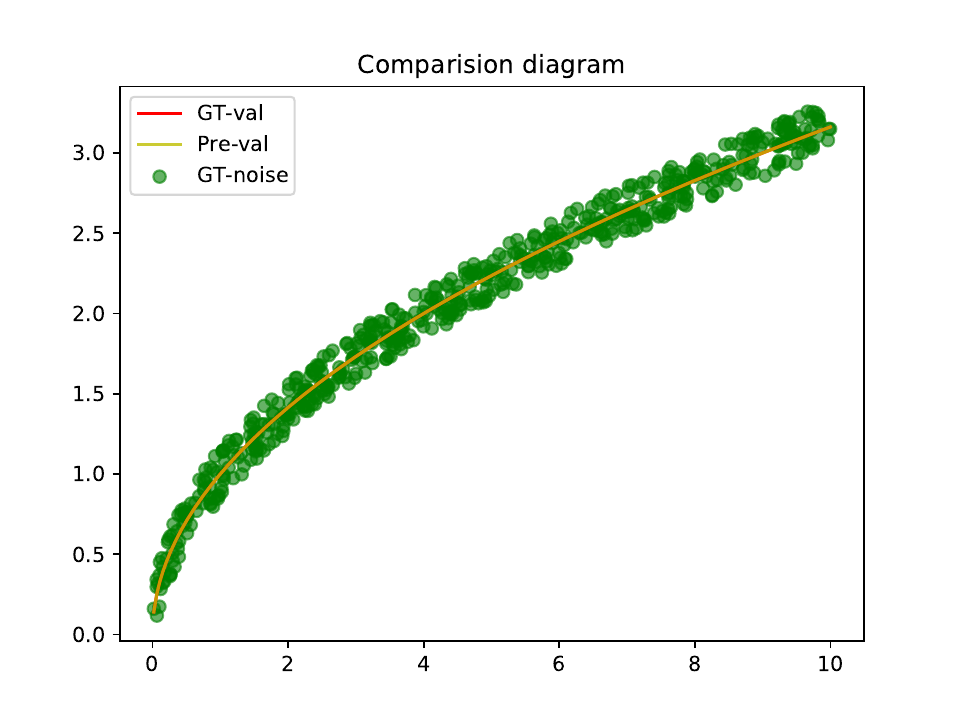}
    \label{1n}}\\
    \subfloat[Nguyen-9-TV]{
       \includegraphics[width=0.11\linewidth]{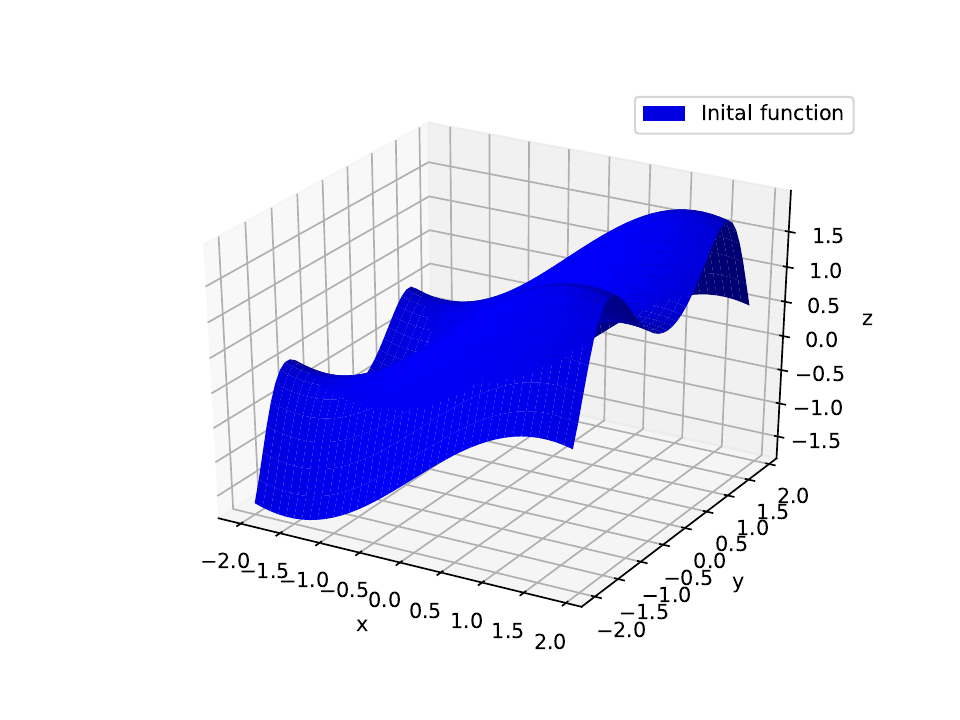}
    \label{1o}}
	  \subfloat[Nguyen-9-PRE]{
        \includegraphics[width=0.11\linewidth]{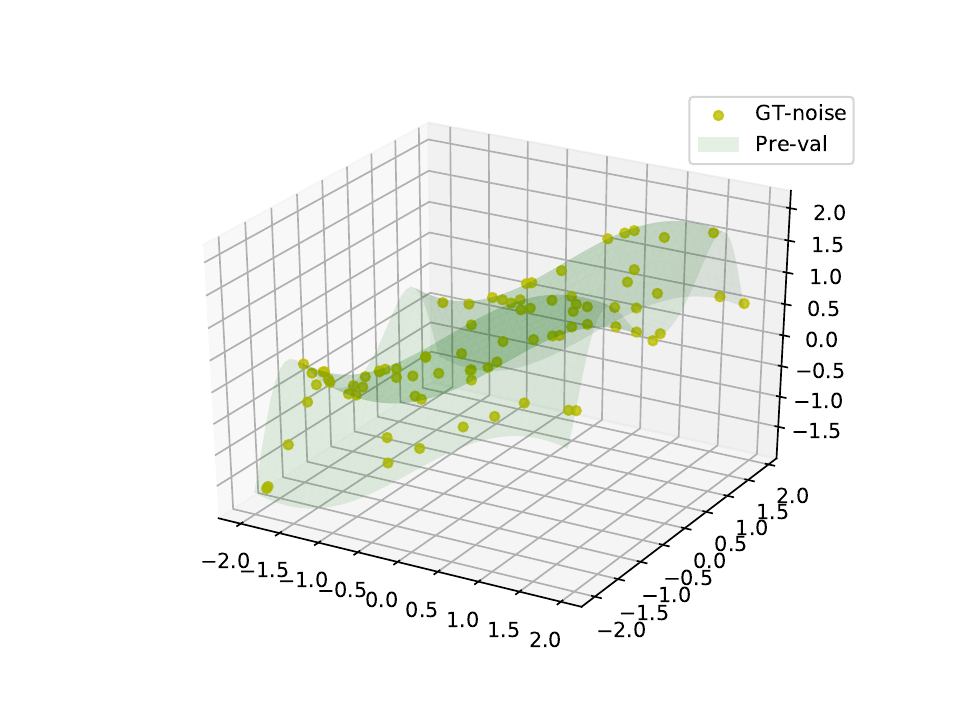}
    \label{1p}}
    \subfloat[Nguyen-10-TV]{
        \includegraphics[width=0.11\linewidth]{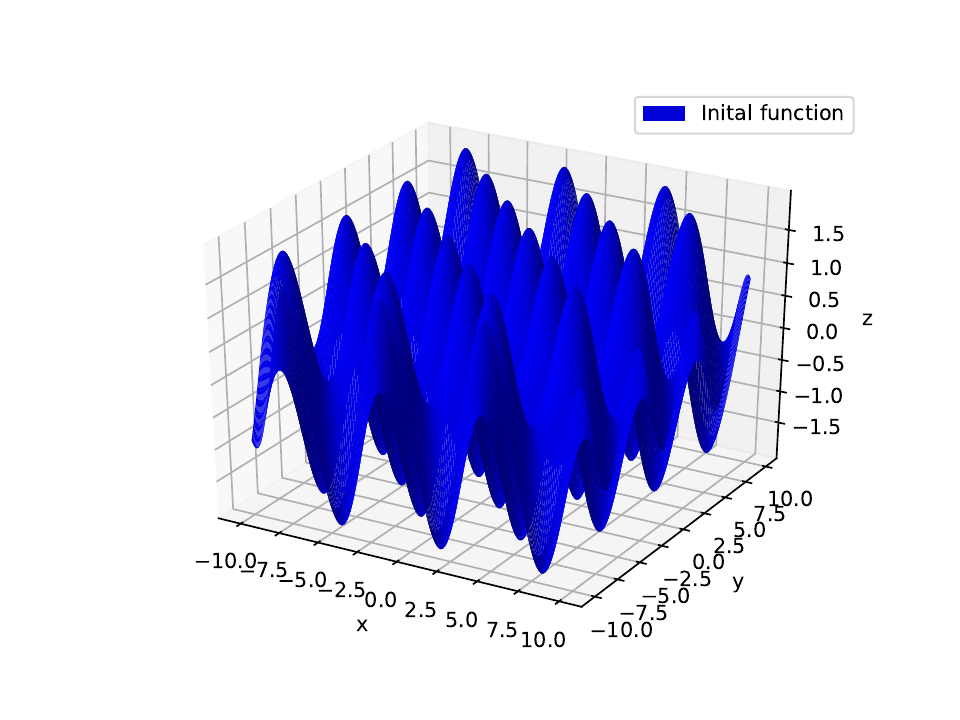}
    \label{1q}}
    \subfloat[Nguyen-10-PRE]{
        \includegraphics[width=0.11\linewidth]{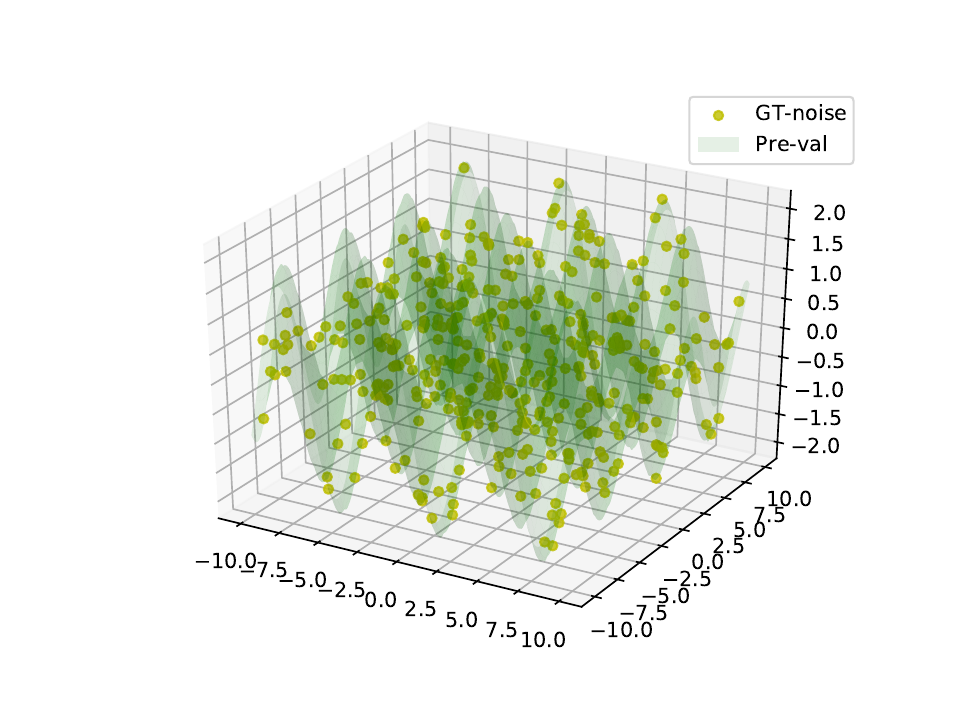}
    \label{1r}}
    \subfloat[Nguyen-11-TV]{
       \includegraphics[width=0.11\linewidth]{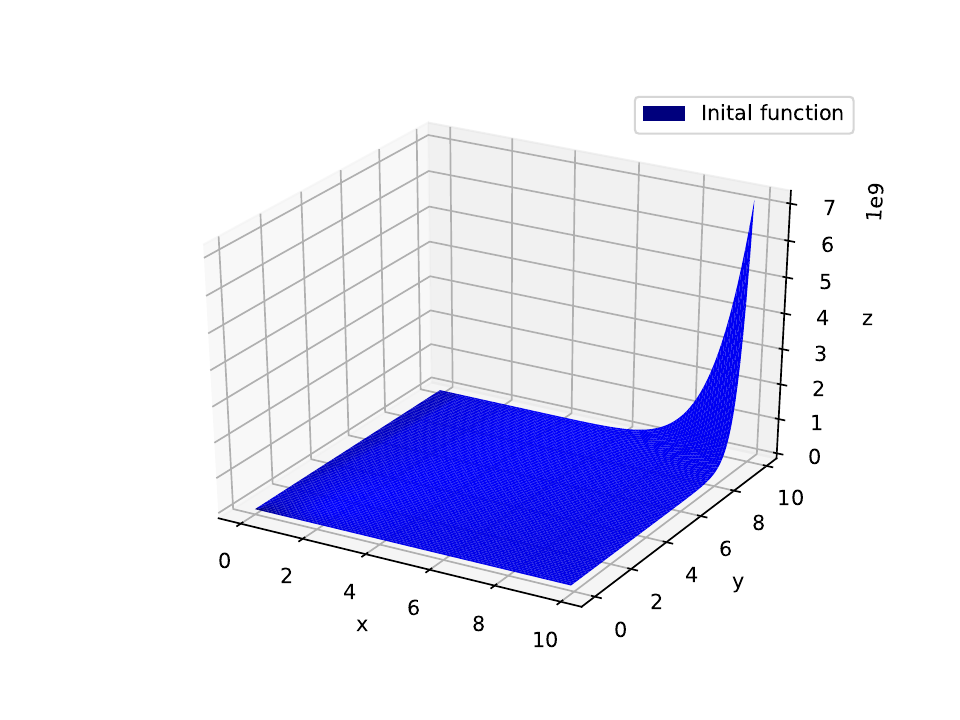}
    \label{1s}}
	  \subfloat[Nguyen-11-PRE]{
        \includegraphics[width=0.11\linewidth]{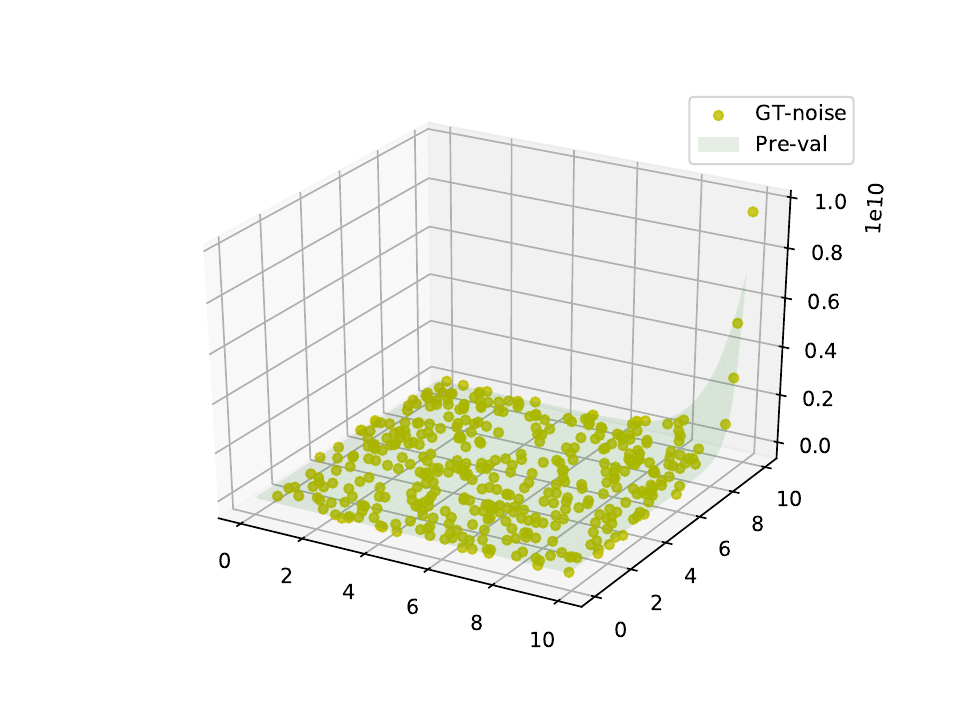}
    \label{1t}}
    \subfloat[Nguyen-12-TV]{
        \includegraphics[width=0.11\linewidth]{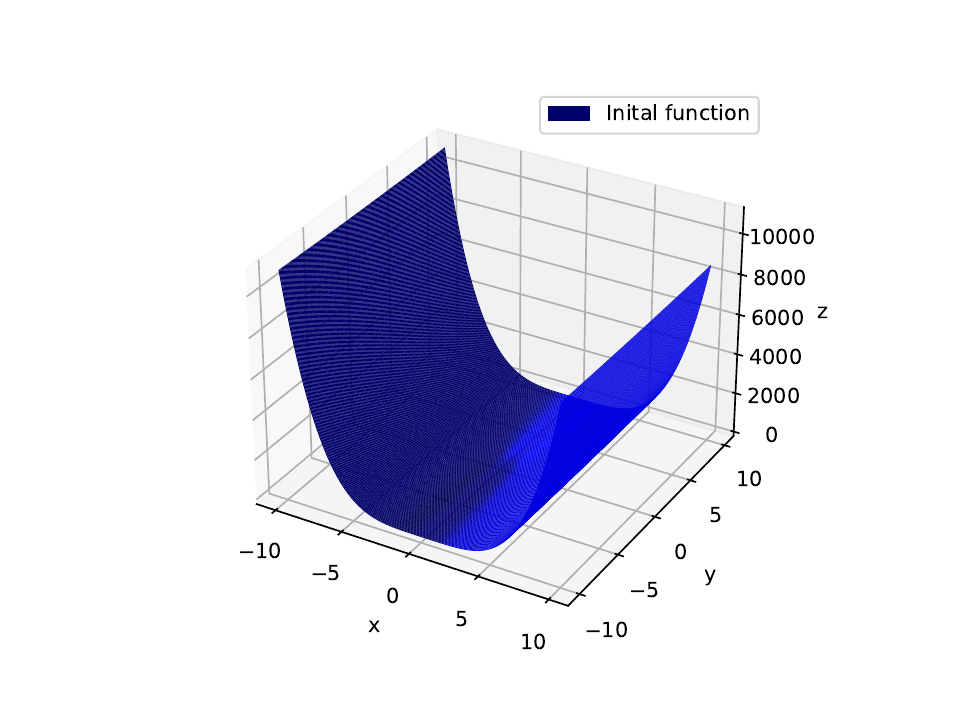}
    \label{1u}}
    \subfloat[Nguyen-12-PRE]{
        \includegraphics[width=0.11\linewidth]{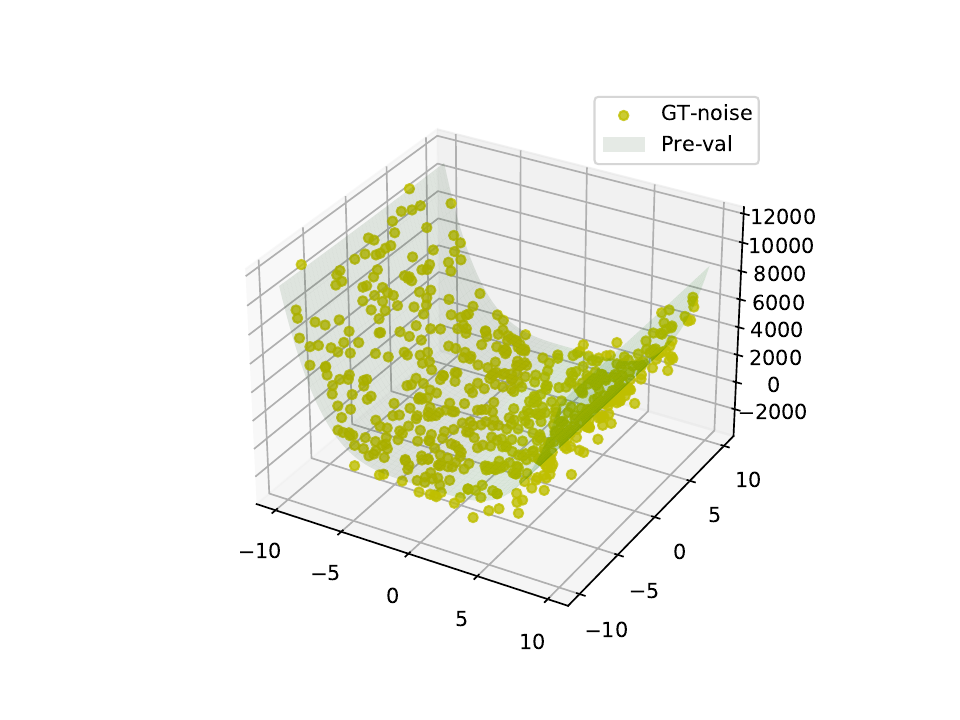}
    \label{1v}}\\
	  \caption{ Figure a and Figure b show the fitting results of Formula \ref{medv1} for Scaled sound pressure level; Figure c displays the correlation coefficient matrix for various variables of Airfoil-self-noise; Figure d and Figure e show the fitting results of Formula \ref{medv2} for Scaled sound pressure level. Figure f displays the fitting results of the Formula\ref{medv2} for global temperature changes; Figures g through n show the fitting results of X-Net on univariate benchmarks. Figures o through v display the prediction outcomes of X-Net on the multivariate benchmark, where `-TV' denotes true values and `-PRE' represents predicted values;
	  }
	  \label{fig4} 
\end{figure*}
\begin{equation}
\label{medv1}
\displaystyle Y = 97.7 + \frac{-x_2 + x_4 + 292}{x_1*x_3*x_5 + 10.5}
\end{equation}

\begin{equation}
\label{medv2}
\displaystyle Y = 130 - \frac{167.37}{\frac{x_4}{x_1*x_3*x_5} + 5.67} 
\end{equation}
The comparison between the predicted data and the real data for the two formulas above is shown in Figure \ref{fig4}. Figure \ref{2a} and Figure \ref{2b} demonstrate the fitting results of Equation \ref{medv1}. Note: To more clearly present the fitting outcomes of Equation \ref{medv1}, without altering the $R^2$ value, we sorted the data points in Figure a based on the predicted values $y_{pre} $ in ascending order, resulting in Figure \ref{2b}. From Figure \ref{2b}, it is evident that Equation \ref{medv1} passes through the center of the data points, robustly fitting the data points. Figure \ref{2d} and Figure \ref{2e} depicts the fitting results of Equation \ref{medv2}, with Figure \ref{2e} undergoing the same data processing as Figure \ref{2b}. From equation \ref{medv1} above, we can observe that the predicted value y is directly proportional to $x_1$, $x_2$, $x_3$, and $x_5$, and inversely proportional to $x_4$. The situation reflected in the formula is consistent with the correlation coefficient heatmap \ref{2c}. From the graph, we can clearly see that only the fourth feature variable is directly proportional to the predicted result $Y$, while the other variables are inversely proportional to the result $Y$. Especially in equation \ref{medv2}, our algorithm only learned from four variables, $x_1$, $x_3$, $x_4$, and $x_5$, and missed variable $x_2$. Although it may not seem like a perfect result, we can see from the correlation heatmap that variable $x_2$ has a high correlation with $x_5$, with a correlation coefficient of 0.75, almost linearly related. Therefore, it is even wiser to retain $x_5$ and discard $x_2$ in the equation. This indirectly reflects the strong learning ability of our algorithm.

\subsubsection{Modeling in environmental science
 (Prediction of Earth's temperature change)\textsuperscript{\cite{earth-temp}}.}
Since the pre-industrial era, human emissions of carbon dioxide (CO2), methane (CH4), and nitrous oxide (N2O) have made significant contributions to global warming. Therefore, exploring the relationship between greenhouse gas emissions and global average temperature changes has become an important goal of international climate research. Here, we applied the historical cumulative emissions of CO2, CH4, and N2O between 1950 and 2022 to predict changes in the global average surface temperature. We used the data from the first 40 years as a training set and the data from the later 32 years as a testing set. The final results obtained from X-Net are as follows.
\begin{equation}
\label{eq27}
\displaystyle Y = 0.000450 * (x_1+x_2+x_3) - 7.61*10^{-8}
\end{equation}
Here, $Y$ represents the global average temperature change, $x_1$ represents the global cumulative CH4 emissions in units of ($PgCO_2-e_{100}$); $x_2$ represents the global cumulative CO2 emissions in units of ($PgCO_2-e_{100}$), and $x_3$ represents the global cumulative N2O emissions in units of ($PgCO_2-e_{100}$). The test results are shown in Figure \ref{2f}. Equation \ref{eq27} clearly shows a direct proportionality between changes in global mean temperature and the sum of emissions from the three greenhouse gases.

\section*{Discussion }\label{sec5}
In this study, we conduct an in-depth exploration of next-generation neural networks and introduce a novel neural network model named X-Net. Theoretically, this model allows the use of any differentiable function as its activation function, which is not static but dynamically learned under the guidance of gradient information. Furthermore, the network structure of X-Net is dynamically adaptable, and capable of self-adjusting at the neuronal level in real-time, including both growth and reduction, to better adapt to specific tasks while minimizing the issues of parameter redundancy and insufficiency. Notably, when dealing with relatively simple tasks, X-Net can derive a simplified and interpretable mathematical formula, which is particularly beneficial for scientific research.

Specifically, we design the network structure as a tree and in order to improve the representation ability of the X-Net, we extend the activation function of the neural network to the activation function library, which not only contains the traditional activation functions such as $[ReLU, sigmoid...]$ but also includes the basic functions such as $[+, -, \times, \div, sin, cos, exp, sqrt, \\log...]$. Next, we propose an alternating backpropagation mechanism, which can optimize not only the parameters of the network but also the activation function of the network nodes and the network structure. In particular, we take the output of each activation function as a variable $ E $, differentiate the output of nodes of each layer through the chain rule, and then update it by the backpropagation algorithm. Finally, we select the activation function of nodes according to the updated $E$. \\
X-Net achieves comparable performance to MLP for both classification and regression. However, in terms of network structure complexity, X-Net is far less than MLP. In addition, to further test X-Net's ability to aid scientific discovery, we tested X-Net in multiple scientific fields, including social science, environmental science, energy science, and space science. In the end, X-Net came up with a concise, analyzable mathematical formula for modeling data in various disciplines and accurately reflected the relationship between the variables $X$ and the predicted value $Y$ in the data.\\
X-Net has great application potential. For example, we can use it to do algorithmic distillation, to distill a complex network into a simple X-Net. Or we could try to replace the fully connected layers in Transformers, GPT, or other deep learning networks and drastically reduce the model size. It can speed up the inference efficiency and reduce the deployment cost of the model. We can also use it to solve partial differential equations(PDE) so that it is possible for us to obtain an analytical solution to the partial differential equation. We can also apply X-Net to healthcare, finance, and other fields with high interpretability requirements to open up the MLP black box system. In summary, we can theoretically use X-Net to do a lot of things that MLP can do.

X-Net also has a number of drawbacks, such as training is also relatively time-consuming. There are even occasional cases where the training fails to converge.

In summary, X-Net provides a new perspective and opens a new possibility for the study of next-generation neural networks. We sincerely hope that X-Net can be a little inspiring to the following researchers.

\section{Methods}\label{sec3}
\begin{figure*} 
    \centering
	  \subfloat[The overall flowchart of X-Net]{
	  \hspace{-1.9cm} 
       \includegraphics[width=1.25\linewidth]{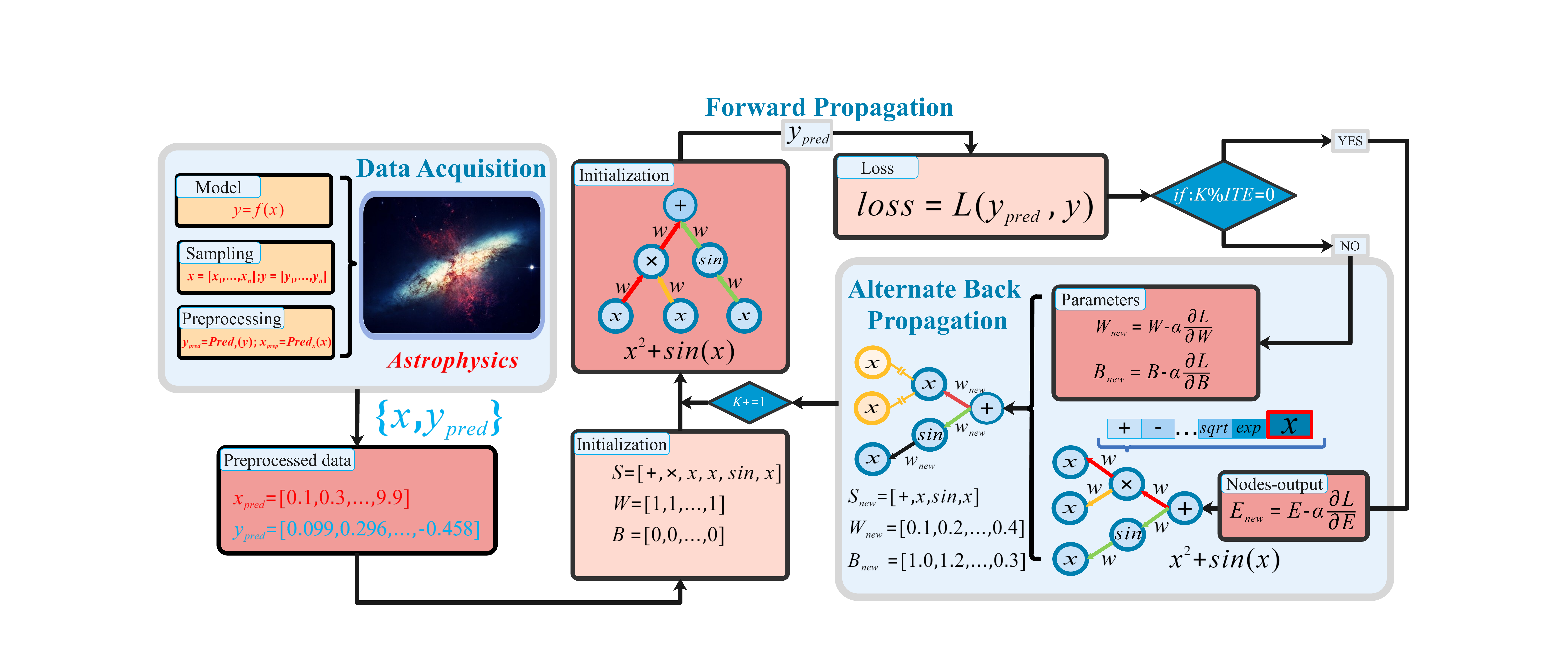}
       \label{3a}}\\
    \subfloat[Comparison between X-Net and MLP structures]{
        \includegraphics[width=0.36\linewidth]{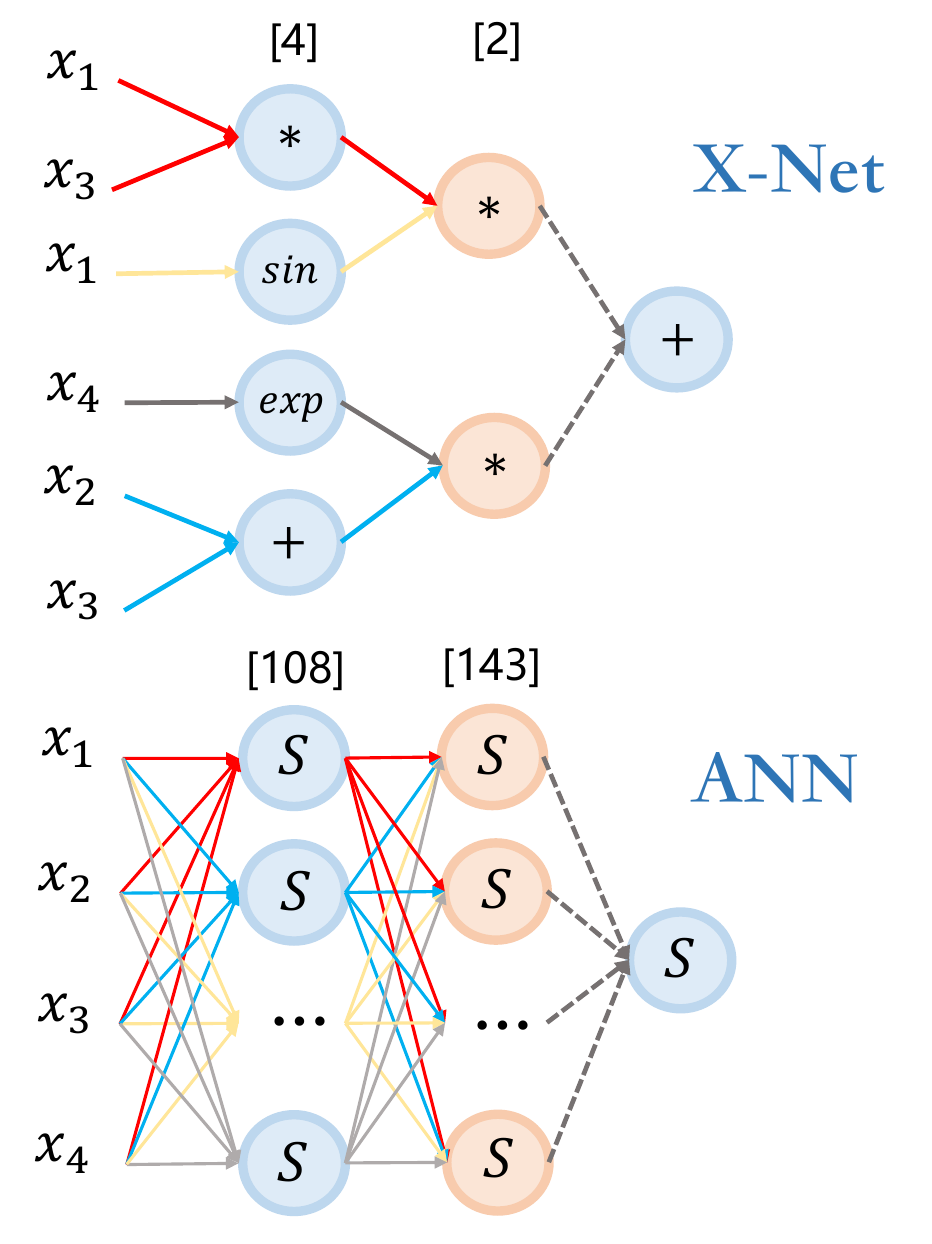}
        \label{3b}}
    \subfloat[Schematic of X-Net forward propagation]{
       \includegraphics[width=0.64\linewidth]{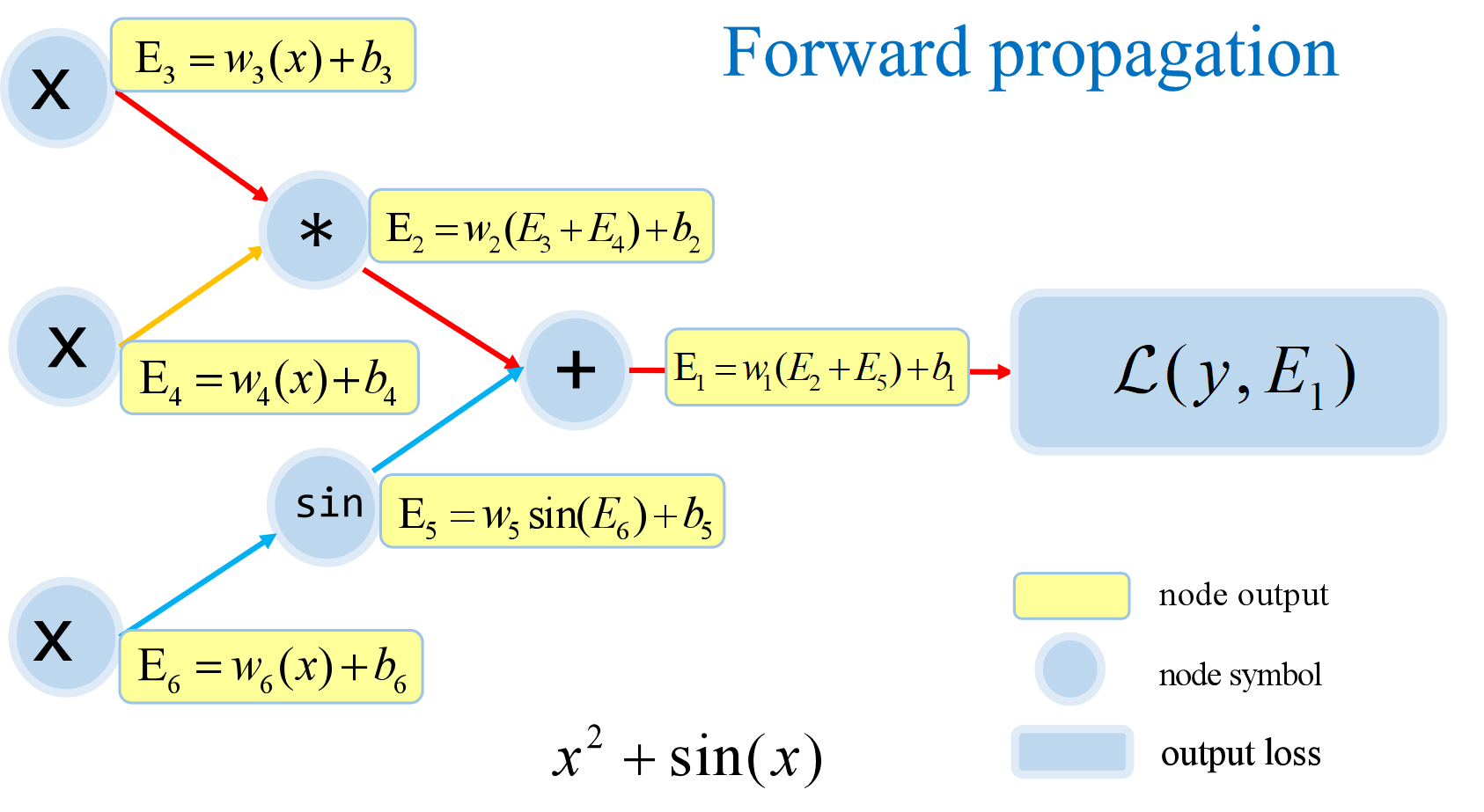}
       \label{3c}}\\
	  \caption{Figure a illustrates the algorithmic flowchart of X-Net; From Figure b, we observe a performance discrepancy between X-Net and MLP in the same regression task, with X-Net having a significantly lower network complexity compared to MLP;
	  Figure c describes the forward propagation process of our algorithm in detail;
   } \label{fig3} 
\end{figure*}

As shown in Fig. \ref{3a}, X-Net is composed of several steps: (1) Feed the training data $\{x,y_{\rm noise}\}$ which are obtained from the sampling of mathematical expressions or observations of real-world to the network; (2) Initial the tree-shape network structure and constants, denote the constants as $W$ (weights) and $B$ (bias); (3) Perform a forward propagation to obtain the predicted value $y_{\rm pre}$; (4) Compute the loss by loss function; (5) Implement the alternating backpropagation algorithm to alternately update parameters $W, B$ as well as the activation function through updating the node output value $E$; (6) Repeat steps (3)-(5) until X-Net exceeds the expected $R^2$ \textsuperscript{\cite{r2}} or reaches the maximum epoch.

X-Net is a tree-shape network and we update the network through the backpropagation algorithm similar to MLP. The distinctions lie in the backpropagation procedure, in which we add an extra step to substitute proper activation functions by the guide of gradients. Fig. \ref{3b} depicts the structure of X-Net and MLP, respectively. Compared with MLP, the connection of X-Net is sparse and the activation functions are diversiform. Fig. \ref{3c} shows the forward propagation of X-Net, the $E_i$ represents the output of $i^{\rm th}$ node, and $w_i$ and $b_i$ are the constants. X-Net optimizes the parameters $W$ and $B$ and the activation functions, which is the key distinction with MLP. Both the network structure and the neuron types are adjusted during the learning process.

\subsection{Forward Propagation}
We use the pre-order traversal of the X-Net to code since it's a tree-shape network and with a unique mapping between the binary tree and pre-order traversal. Assume the pre-order traversal of a X-Net is $S=[s_1,s_2,...,s_m]$, where $m$ is the number of neuron or number of node, $s_i$ indicates the $i^{\rm th}$ neuron. The activation function of each $s_i$ is selected from a library $\{ReLu, sigmoid, sin, cos, log, exp, sqrt,..., +,-,\times,\div,x_1,x_2,...\}$ that contains unary, binary, and variables. X-Net is initiated according to the arity of the activation function of each neuron: if the activation function requires two inputs, such as $+,-,\times,\div$, then the neuron has two inputs; otherwise, the neuron has one input. Each neuron will have two distinct constant values $w_i$ and $b_i$. The forward propagation is conducted from left to right (as shown in Fig. \ref{3c}), or alternately, from leaf to root for the binary tree. The root node outputs the predicted value $\hat{y}$. Finally, the loss function is calculated according to different types of tasks.

\subsection{Alternating Backpropagation}
While traditional backpropagation optimizes a network's parameters, X-Net requires updating the parameters and activation functions. Therefore, we designed the alternating backpropagation algorithm to update X-Net's parameters and activation functions alternately. In the process of alternating backpropagation, each neuron is viewed as a variable $E_i=f_i(E_{\rm left}^i, E_{\rm right}^i)$ requiring optimization, where $E_i$ is the output of the $i^{\rm th}$ node, $E_{\rm left}^i$ and $E_{\rm right}^i$ are the left and right inputs of the node $i$, respectively. $f_i$ is the numerical computation corresponding to the node activation function. We calculate the derivative of $E_{\rm left}^i$ and $E_{\rm right}^i$ by the chain rule \textsuperscript{\cite{cr}} and update them with the stochastic gradient descent (SGD) \textsuperscript{\cite{sgd1,sgd2}}, or other numerical optimization algorithms such as BFGS, L-BFGS. After the gradients have been updated, the activation functions are selected on the basis of $E^i_{\rm new}$ (details of selection and substitution are described in section \ref{select_activation_function} and \ref{substitute_of_activate_function}, respectively), then the constants in the network are updated by SGD. Through the alternating backpropagation, we obtain the gradients of each node's output $E_i$ and its corresponding constants $w_i$, $b_i$.

Take a case as an example, as shown in Fig. \ref{3c}, the node $E_1=w_1f_1(E_2, E_5)+b_1$, where $f_1$ is the addition function, the gradients of $E_1$, $w_1$, and $b_1$ are:
\begin{align}
\displaystyle
    \nabla E_1 &= \frac{\partial \mathcal{L}}{\partial E_1} \vspace{3ex}\\
    \nabla w_1 &= \frac{\partial \mathcal{L}}{\partial E_1}\frac{\partial E_1}{\partial w_1} =\nabla E_1\frac{\partial E_1}{\partial w_1}\vspace{3ex}\\
    \nabla b_1 &= \frac{\partial \mathcal{L}}{\partial E_1}\frac{\partial E_1}{\partial b_1} =\nabla E_1\frac{\partial E_1}{\partial b_1} = \nabla E_1\vspace{3ex}
\end{align}
Where $\mathcal{L}$ is the loss function. The $E_1$ and constants are updated by the following formulas:
\begin{align}
\displaystyle
    E_{1_{\rm new}} = E_1-\alpha \nabla E_1 \\
    w_{1_{\rm new}}=w_1-\alpha \nabla w_1 \\
    b_{1_{\rm new}}=b_1-\alpha \nabla b_1 
\end{align}
Where $\alpha$ is the learning rate. According to the node chain rule, the gradients of $E_2$, $E_5$ are:
\begin{align}
\displaystyle
    \nabla E_2 &= \frac{\partial \mathcal{L}}{\partial E_2}=\frac{\partial \mathcal{L}}{\partial E_1}\frac{\partial E_1}{\partial E_2}=\nabla E_1\frac{\partial E_1}{\partial E_2}\vspace{3ex}\\
    \nabla E_5 &= \frac{\partial \mathcal{L}}{\partial E_5}=\frac{\partial \mathcal{L}}{\partial E_1}\frac{\partial E_1}{\partial E_2}\frac{\partial E_2}{\partial E_5}=\nabla E_2\frac{\partial E_2}{\partial E_5} \vspace{3ex}
\end{align}

For node 2, 5, $E_2=w_2(E3+E_4)+b_2$, $E_5=w_5\sin(E_6)+b_5$. $w_2$, $b_2$ are related to $E_2$, $W_5$, $b_5$ are related to $E_5$, thus, the gradients are calculated as:
\begin{align}
\displaystyle
    \nabla w_2 &=\frac{\partial \mathcal{L}}{\partial w_2}=\frac{\partial \mathcal{L}}{\partial E_1}\frac{\partial E_1}{\partial E_2}\frac{\partial E_2}{\partial w_2}=\nabla E_2\frac{\partial E_2}{\partial w_2} \vspace{3ex}\\
    \nabla b_2 &= \frac{\partial \mathcal{L}}{\partial b_2}=\frac{\partial \mathcal{L}}{\partial E_1}\frac{\partial E_1}{\partial E_2}\frac{\partial E_2}{\partial b_2}=\nabla E_2\frac{\partial E_2}{\partial b_2}=\nabla E_2 \vspace{3ex}\\
    \nabla w_5 &=\frac{\partial \mathcal{L}}{\partial w_5}=\frac{\partial \mathcal{L}}{\partial E_1}\frac{\partial E_1}{\partial E_2}\frac{\partial E_2}{\partial E_5}\frac{\partial E_5}{\partial w_5}=\nabla E_5\frac{\partial E_5}{\partial w_5} \vspace{3ex}\\
    \nabla b_5 &=\frac{\partial \mathcal{L}}{\partial b_5}=\frac{\partial \mathcal{L}}{\partial E_1}\frac{\partial E_1}{\partial E_2}\frac{\partial E_2}{\partial E_5}\frac{\partial E_5}{\partial b_5}=\nabla E_5\frac{\partial E_5}{\partial b_5}=\nabla E_5 \vspace{3ex}
\end{align}

Then, we update the values of the constants by subtracting the product of the learning rate 
 and gradient.

\subsection{Evaluate Activation Function}
\label{select_activation_function}
The updated output $E_{\rm new}$ of each neuron is arranged according to the pre-order traversal: $E_{\rm new}=[E_{1_{\rm new}}, E_{2_{\rm new}},..., E_{m_{\rm new}}]$. We aim to select the activation from the library that makes the prediction more accurate. For the $i^{\rm th}$ node, $E_i=f_i(E^i_{\rm left_{old}}, E^i_{\rm right_{old}})$, where $f_i$ is the addition function, $E^i_{\rm left_{old}}$ and $E^i_{\rm right_{old}}$ are the outputs of the left and right node of the $i^{\rm th}$ node, respectively. Suppose the values of the outputs are updated to $E_{i_{\rm new}}$, $E^i_{\rm left_{new}}$, and $E^i_{\rm left_{new}}$. The $E^i_{\rm left_{new}}$ and $E^i_{\rm left_{new}}$ are fed to the activation function library to calculate the new output of node $i$: $E_{i,j}'=f_i(E^i_{\rm left_{new}}, E^i_{\rm left_{new}})$, where $f_i\in \{+,-,\times,\div, sin, cos, log, sqrt,exp, relu, sigmoid,x\}$, $E_{i}'\in\mathbb{R}^{12}$, $j=1,2,...,12$ is the index of the activation function library. Then we calculate the difference between $E_{i.j}'$ and $E_{i_{\rm new}}$ and take the absolute value resulting in $G=\{g_1,g_2,...,g_{12}\}$, where $g_j=|E^i-E_{i_{\rm new}}|$. We select the activation function that has the smallest $g_{\rm min}$. To ensure steady training, we set a threshold (e.g. 0.01), $g_{\rm min}$ must lower than the threshold.

\subsection{Substitute Activation Function}
\label{substitute_of_activate_function}
X-Net has diversified activation functions, and the exchange of activation functions brings difficulties to the forward propagation. For example, the arity of an activation function may change from one to two, the leaf node can not have a child node, etc. Therefore, we design several rules to ensure the X-Net can successfully conduct forward propagation after the activation functions have been updated, Fig. \ref{fig4} shows the different types of substitution. Specifically, We set five rules:
\begin{itemize}
    \item[(1)] If the arity of the activation function does not change, substitute the old activation function to the new;
    \item[(2)] If the activation function of a node changes from unary to binary, keep the left child and add a leaf node $x$ as the right child;
    \item[(3)] If the activation function of a node changes from binary to unary, keep the child that has better fitting results and drop the other;
    \item[(4)] If the activation function of a node changes to the leaf node, it does not have child nodes;
    \item[(5)] If a leaf node changes to a unary node, the left child is added as the input of the node; If a leaf node changes to a binary node, the left and right children are added to the node.
\end{itemize}

\subsection{Solutions for numerical computation}
To avoid the gradient explosion problem, we revise some computation rules: (1) The outputs of all nodes are truncated during the computation process to prevent numerical overflow, specifically, $-V\leq E_i \leq V$, if $|E_i|\geq V$, $E_i=\frac{E_i}{|E_i|}V$; (2) We apply Gradient Clipping and limit the gradient in $[23,24]$. Assume the gradient value is $g_i$, which lies in the range $[-G_{max}, G_{min}]\cup [G_{min}, G_{max}]$ if $|g_i|\geq G_{max}$, then $g_i$ is set as $\frac{g_i}{|g_i|}G_{max}$; if $|g_i|\leq G_{min}$, then $g_i\frac{g_i}{|g_i|}G_{min}$; (3) We select the activation function in the basis of domain of definition, take the $\log(x)$ as an example, the $x$ must satisfy $x>0$.

\subsection{Avoid Getting Stuck in Local Optimum}
The network opts to fall into a local optimum in the process of optimization and training if no additional conditions are applied \textsuperscript{\cite{lo}}. To prevent this problem from scratch, we set a count value $count$ as an indicator. If the loss remains unchanged or increases, $count=count+1$; If $count$ reaches a set threshold, we randomly select a node in X-Net and substitute a randomly selected activation function from the library. Through this ``mutation'' operation, X-Net is more likely to jump out of the local optimum. In our experiments, we found it was effective when the threshold was 20.

\subsection{Adjust Learning Rate by Adaption Function: Ada-$\alpha$}
The learning rate has a huge impact on the converging speed of the network and the search for the optimal solution \textsuperscript{\cite{st1,st2,st3}} in the training process. Therefore, we design an adaption function Ada-$\alpha$ which can adaptively and dynamically adjust the learning rate to achieve better fitting performance. Specifically, Ada-$\alpha$ is calculated by:
\begin{equation}
\label{ada}
\displaystyle\alpha = \frac{\tanh(e^{- |\mathcal{L}_{\rm pre}-\mathcal{L}_{\rm cur}|})}{a}.
\end{equation}
Where $\mathcal{L}_{\rm pre}$ and $\mathcal{L}_{\rm cur}$ represent the loss value in the previous iteration and current iteration, respectively. $a$ is the hyperparameter to adjust the range of learning rate. 

\section*{Acknowledgments}

\paragraph{Funding:}
This work was supported in part by the National Natural Science Foundation of China under Grant 92370117, in part by CAS Project for Young Scientists in Basic Research under Grant YSBR-090, and in part by the Key Research Program of the Chinese Academy of Sciences under Grant XDPB22.
\paragraph{Authors contributions:}
In this work, W.J. Li acted as the corresponding author, coordinating the research activities, injecting creativity into the project, and significantly enhancing its impact and scope. Y. Li was responsible for writing the paper, algorithm conception and design, and related experiments. L. Yu gave detailed guidance on the whole process of paper writing and experimental setup. M. Wu and L. Sun gave many useful suggestions for the design of the algorithm. J. Liu set up the structure of the article in detail. And carefully polished and improved the method part of the article. W.Q. Li and Y. Li collaborated on the writing of the abstract and introduction sections of the article. M. Hao, S. Wei, and Y. Deng helped the authors with some experiments. The completion of this work is inseparable from the efforts of each of the above. Thank you all for your help.
\paragraph{Competing interests:}
All authors of the article have no competing interests


\bibliography{scibib}
\bibliographystyle{unsrt}
  
\newpage
\section*{Supplementary materials}
\section{Pseudocode for X-Net}\label{secA1}
This is the pseudocode for the overall flow of our algorithm. We first initialize a tree neural network, and initialize the activation function $s_i$ for each node, and the parameters $[ \mathcal{W},\mathcal{B}]$ of the network. Then we forward to compute the output of each neuron E. Next we compute the loss, and determine if the current loss on the validation set is due to the previous loss, if yes, we save the current network architecture and the optimal $R^2$. Train the network and repeat until a predetermined number of iterations is reached.
\newcommand{\invtriangle}{\mathbin{\rotatebox[origin=c]{180}{$\triangle$}}}

\begin{algorithm}[H]
\caption{Pseudocode for X-Net.}\label{pseudocode-X-Net} 
\begin{algorithmic}[1]
\State Initialize: $\mathcal{S}=[s_1,s_2,...,s_m];\ \mathcal{W}=[w_1,w_2,...,w_m];\ \mathcal{B}=[b_1,b_2,...,b_m];\ \mathcal{X}=[x_1,x_2,...,x_n];\ \mathcal{Y}=[y_1,y_2,...,y_n]$
\For {$j = 0$ to $ite$}
    \State $\mathcal{S} \leftarrow \mathcal{S}_{best}$
    \State $\mathcal{W} \leftarrow \mathcal{W}_{best}$
    \State $\mathcal{B} \leftarrow \mathcal{B}_{best}$
    \For {$i = 0$ to $n$}
        \For {$k = 0$ to $100$}
            \State The nodes: $\mathcal{A}=[a_0,a_1,...,a_m]$ \Comment{Initializing nodes.}
            \State Constructing a tree-like network: $\mathcal{A} \leftarrow \mathcal{S}$
            \For {$k = 0$ to $m$}
                \State $\mathcal{A}_k \leftarrow \mathcal{S}_K$\Comment{Each node is assigned a specific activation function}
            \EndFor
            \State $E=[E_0,E_1,...,E_m]$ \Comment{Initialize each node output}
            \For {$k = 0$ to $m$}
                \State $E_k \leftarrow w_k * S_k(x_l, x_r) + b_k$ \Comment{Compute the output of each node.}
            \EndFor
            \State $\hat{y} \leftarrow E_0$
            \State loss $\leftarrow \mathcal{L}(y_i, \hat{y})$ \Comment{Calculating the loss.}
            \State $E_{val} = f(X_{val})$
            \State $\mathcal{R}^2 = r2(E_{val},\mathcal{Y}_{val})$\Comment{Get the $R^2$ of the current expression on the validation set.}
            \State SaveBest($\mathcal{S},\mathcal{R}^2$)
            \State Train($key,ITE,E,\mathcal{W},\mathcal{B}$)
        \EndFor
    \EndFor
\EndFor
\end{algorithmic}
\end{algorithm}

\section{Pseudocode for SaveBest}\label{pseudocode-SaveBest}
This pseudocode shows that in a regression task if the current $R^2$ is greater than the previous best $R^2$, we save the optimal $R^2$, parameters, and activation function groups (in order of network preorder traversal).
\begin{algorithm}
\caption{SaveBest}\label{alg:alg1}
\begin{algorithmic}[1] 
\State Variables: $R^2$; $R_{\text{best}}$
\State Start:
\If{$R^2 \geq R_{\text{best}}$}
  \State $\mathcal{R}_{\text{best}} \leftarrow R^2$ \Comment{Save the best $R^2$.}
  \State $\mathcal{S}_{\text{best}} \leftarrow \mathcal{S}$ \Comment{Save the best group of activation functions.}
  \State $\mathcal{W}_{\text{best}} \leftarrow \mathcal{W}$\Comment{Save the best $\mathcal{W}$.}
  \State $\mathcal{B}_{\text{best}} \leftarrow \mathcal{B}$\Comment{Save the best $\mathcal{B}$.}
\EndIf
\end{algorithmic}
\end{algorithm}
\section{Pseudocode for Train}\label{pseudocode-train}
This pseudocode shows the detailed training process of X-Net. We use alternating backward transmission to optimize parameters $[W, B]$ and node output $E$ alternately. We use the hyperparameters $key$ and $ITE$ to control how often $[W, B]$ and $E$ are optimized.

\begin{algorithm}[H]
\caption{Train}\label{alg:alg1}
\begin{algorithmic}
\State Variables: $key$; $ITE$; $E=[E_1,E_2,...,E_m]$;
\State \hspace{\algorithmicindent} $\mathcal{W}=[w_1,w_2,...,w_m]$, $\mathcal{B}=[b_1,b_2,...,b_m]$
\If{$key \% ITE \neq 0$} \Comment{If the condition is not satisfied, train and update the $\mathcal{W}$, $\mathcal{B}$}
    \State $\mathcal{W}_{new}[k] \leftarrow \mathcal{W}[k] - \alpha \frac{\partial L}{\partial \mathcal{W}[k]}$
    \State $\mathcal{B}_{new}[k] \leftarrow \mathcal{B}[k] - \alpha \frac{\partial L}{\partial \mathcal{B}[k]}$
    \State $key$ += 1 
\EndIf
\If{$key \% ITE = 0$}\Comment{If the condition is not satisfied, train and update the outputs of nodes.}
    \For{$k = 0$ to $m$}
        \State $E_{new}[k] \leftarrow E[k] - \alpha \frac{\partial L}{\partial E[k]}$
    \EndFor
    \State UpdateSymbols($E$)\Comment{Update the activate functions.}
    \State $key$ += 1
\EndIf
\end{algorithmic}
\end{algorithm}
\section{Pseudocode for SaveBest}\label{pseudocode-UpdateSymbols}
This pseudocode shows the process of node activation function selection for X-Net. Specifically, we feed the updated input of each node (the output of the child node) into all the candidate activation functions for numerical calculation to obtain $\mathcal{SY}$. Then the activation function corresponding to the closest value in $\mathcal{SY}$ to the updated output of the current node is selected as the activation function of the current node.
\begin{algorithm}[H]
\caption{UpdateSymbols}\label{alg:alg1}
\begin{algorithmic}[1]
\State Variables: $E_{new}=[E_{1new},E_{2new},...,E_{mnew}]$;
\State $activation\ functions = [+,-,\times,\div, \sin, \cos, \sqrt, \log, \exp, sigmoid, relu, x]$
\For{$k = 0$ to $m$}
    \State $\mathcal{SY} \leftarrow \{E[k]_{newl}+E[k]_{newr}, E[k]_{newl}-E[k]_{newr},$
    \State \hspace{\algorithmicindent} $E[k]_{newl} \times E[k]_{newr}, E[k]_{newl} \div E[k]_{newr},$
    \State \hspace{\algorithmicindent} $sin(E[k]_{newl}), cos(E[k]_{newl}), sqrt(E[k]_{newl}),$ 
    \State \hspace{\algorithmicindent} $log(E[k]_{newl}), exp(E[k]_{newl}), sigmoid(E[k]_{newl}),$
    \State \hspace{\algorithmicindent} $relu(E[k]_{newl}), x\}$\Comment{The new input of this node is fed into the candidate activation function for calculation.}
    \State $Choice \leftarrow |\mathcal{SY} - E[k]_{new}|$
    \State $Index \leftarrow \operatorname{arg\,min}(Choice)$\Comment{Select the new activation function with that section.}
    \State $S[k] \leftarrow activation\ functions[Index]$ \Comment{Activation function replacement.}
\EndFor
\end{algorithmic}
\end{algorithm}

\newpage
\section{Details of the Nguyen dataset} 
We evaluated X-Net and MLP on the Nguyen symbolic regression benchmark suite \textsuperscript{\cite{uy2011semantically}}, comprising twelve benchmark expressions widely recognized and utilized within the symbolic regression field \textsuperscript{\cite{white2013better}}.
Each benchmark is defined by a ground truth expression, described in Table \ref{Nguyen}. The curves of these formulas have both high frequency and low frequency, simple and complex, which can better reflect the fitting ability of the algorithm.

\begin{table*}[ht]
\centering
\caption{ Symbol library and value range of the three data sets Nguyen, Korns, and Jin. 
}
\begin{footnotesize}
\begin{tabular}{cc}
\toprule[1.45pt]
\toprule
\textbf{Name} & \textbf{Expression}  \\ \hline
Nguyen-1 & $x_1^3+x_1^2+x_1$ \\
Nguyen-2 & $x_1^4+x_1^3+x_1^2+x_1$  \\
Nguyen-3 & $x_1^5+x_1^4+x_1^3+x_1^2+x_1$ \\
Nguyen-4 & $x_1^6+x_1^5+x_1^4+x_1^3+x_1^2+x_1$ \\
Nguyen-5 & $\sin(x_1^2)\cos(x)-1$\\
Nguyen-6 & $\sin(x_1)+\sin(x_1+x_1^2)$ \\
Nguyen-7 & $\log(x_1+1)+\log(x_1^2+1)$ \\
Nguyen-8 & $\sqrt{x}$  \\
Nguyen-9 & $\sin(x)+\sin(x_2^2)$  \\
Nguyen-10 & $2\sin(x)\cos(x_2)$  \\
Nguyen-11 & $x_1^{x_2}$ \\
Nguyen-12 & $x_1^4-x_1^3+\frac{1}{2}x_2^2-x_2$ \\
\hline
\end{tabular}
\end{footnotesize}
\label{Nguyen}
\end{table*}

\section{Small Sample learning ability}

MLP training often requires a large number of training samples. However, in real life, the cost of sample collection is very high in many cases, and it is difficult for us to obtain many samples. Therefore, the ability of the model to learn with a small number of samples is very important. To test the small sample learning ability of X-Net and MLP. We tested it on the Nguyen dataset. Specifically, we first sample 10,000 points for each test data. We then sample 50,100,500, and 1,000 points from these. These points are then used to train X-Net and MLP, respectively. Whenever the training is complete, we use the remaining points as the test set to test the model. The specific results are shown in table \ref{tab_small}.

\begin{table*}[t]
\center
\caption{Comparison of Small Sample learning ability.
\label{tab_small}}
\resizebox{\textwidth}{36mm}{
\begin{tabular}{ccccccccccccccccc}
\toprule[1.45pt]
\toprule[1pt]
Baselines& \multicolumn{8}{c}{X-Net}&\multicolumn{8}{c}{MLP}\\
\cmidrule(lr){2-9}
\cmidrule(lr){10-17}
Points& \multicolumn{2}{c}{50}&\multicolumn{2}{c}{100}&\multicolumn{2}{c}{500}&\multicolumn{2}{c}{1000}&\multicolumn{2}{c}{50}&\multicolumn{2}{c}{100}&\multicolumn{2}{c}{500}&\multicolumn{2}{c}{1000}\\  
\cmidrule(lr){2-9}
\cmidrule(lr){10-17}
& Train & test  & Train& test & Train& test& Train& test& Train& test& Train& test& Train& test& Train& test\\ 
\cmidrule(lr){2-3}
\cmidrule(lr){4-5}
\cmidrule(lr){6-7}
\cmidrule(lr){8-9}
\cmidrule(lr){10-11}
\cmidrule(lr){12-13}
\cmidrule(lr){14-15}
\cmidrule(lr){16-17}
Nguyen-1 &  $1.00 $ & $1.00$  & $1.00$&$ 1.00 $ &$ 1.00 $&$1.00$ &$ 1.00 $&$1.00$&$ 0.99$ &$ 0.99$&$ 0.99$ &$ 0.99$ &$ 0.99 $&$0.99$ &$ 0.99$ &$ 0.99$\\
Nguyen-2 & $ 1.00$  & $1.00$ &$ 1.00$ &$ 1.00$ &$ 1.00 $&$1.00$&$ 1.0$ & $ 1.0$ &$ 0.99$ &$ 0.99$&$ 0.99$ &$ 0.99$&$ 0.99 $&$0.99$&$ 0.99$ &$ 0.99$\\
Nguyen-3 & $ 0.99$  &$ 0.99 $  &$0.99$ &$ 0.99 $&$ 0.99 $&$0.99$&$ 0.99 $&$ 0.99$ &$ 0.99$ &$ 0.98$&$ 0.99$ &$ 0.99$&$ 0.99 $&$0.99$&$ 0.99$ &$ 0.99$\\
Nguyen-4 & $ 0.99$  &$ 0.99 $  &$0.99$ &$ 0.99 $&$ 0.99 $&$0.99$&$ 0.99 $&$ 0.99$ &$ 0.99$ &$ 0.98$&$ 0.99$ &$ 0.99$&$ 0.99 $&$0.99$&$ 0.99$ &$ 0.99$\\
Nguyen-5 &$  0.99 $ &$ 0.91$ &$0.99$  &$0.98 $ &$ 0.99$&$0.99$&$ 0.99$&$ 0.99$ &$ 0.99$ &$ 0.29$&$ 0.99$ &$ 0.75$&$ 0.99 $&$0.99$&$ 0.99$ &$ 0.99$\\
Nguyen-6 & $  0.99$ &$ 0.99 $ &$0.99 $ &$0.99 $ &$ 0.99 $&$0.99$&$ 0.99 $&$ 0.99$  &$ 0.99$ &$ 0.25$&$ 0.99$ &$ 0.86$&$ 0.99 $&$0.99$&$ 0.99$ &$ 0.99$\\
Nguyen-7 & $  0.99$ &$ 0.99 $ &$0.99 $ &$0.99 $ &$ 0.99 $&$0.99$&$ 0.99 $&$ 0.99$  &$ 0.99$ &$ 0.99$&$ 0.99$ &$ 0.99$&$ 0.99 $&$0.99$&$ 0.99$ &$ 0.99$\\
Nguyen-8 &$  1.00$  &$ 1.00 $ &$ 1.00 $&$1.00$  &$ 1.00 $&$1.00$  &$1.00$ &$ 1.00$ &$ 0.99$ &$ 0.99$ &$ 0.99$ &$ 0.99$&$ 0.99 $&$0.99$&$ 0.99$ &$ 0.99$\\
Nguyen-9 &$  1.00$  &$ 1.00$  &$ 1.00$ &$ 1.00$&$ 1.00 $&$1.00$&$ 1.00$ &$ 1.00$ &$ 0.99 $&$-1.02$  &$ 0.99$ &$ -1.08$ &$ 0.99 $&$0.29$&$ 0.99$ &$ 0.99$\\
Nguyen-10 & $1.00$  & $1.00$ &$1.00$. &$1.00$&$ 1.00 $&$1.00$ & $1.00$ & $1.00$  &$0.99$ &$ 0.12$ &$ 0.99$ &$ 0.89$&$ 0.99 $&$0.99$&$ 0.99$ &$ 0.99$\\
Nguyen-11 & $1.00 $& $1.00 $& $1.00$&$ 1.00 $&$1.00$&$1.00$ &$ 1.00$ & $1.00$  &$ 0.99$ &$ 0.70$ &$ 0.99$ &$ 0.93$&$ 0.99 $&$0.99$&$ 0.99$ &$ 0.99$\\
Nguyen-12 & $0.99$ & $0.97$& $0.99$&$0.99 $&$ 0.99 $&$0.99$& $0.99 $&$ 0.99$ &$ 0.99$ &$ 0.79$ &$ 0.99$ &$ 0.84$&$ 0.99 $&$0.95$&$ 0.99$ &$ 0.98$\\ 
\cmidrule(lr){2-3}
\cmidrule(lr){4-5}
\cmidrule(lr){6-7}
\cmidrule(lr){8-9}
\cmidrule(lr){10-11}
\cmidrule(lr){12-13}
\cmidrule(lr){14-15}
\cmidrule(lr){16-17}
Average & 0.995 & 0.987 & 0.995 & 0.994& 0.995&0.995 &0.995 &0.995&0.990 &0.588 & 0.990&0.761 & 0.990& 0.928& 0.990& 0.989\\
\toprule
\end{tabular}
}
\end{table*}

\section{Complementary Experiment Results}  
We compare X-Net with two methods: Kronecker Neural Networks (Kronecker NNs) and Shinichi\textsuperscript{\cite{shirakawa2018dynamic}}. The brief descriptions of the two methods are:

\begin{itemize}
\item \textbf{Kronecker NN}. 
Kronecker NN is a new neural network framework using Kronecker products for efficient wide networks with fewer parameters. Kronecker NN also introduces the Rowdy activation function, adding trainable sinusoidal fluctuations for better data fitting.

\item \textbf{Shinichi}\textsuperscript{\cite{shirakawa2018dynamic}}. This method generates the probability distribution of the network structure and then optimizes the parameters of the distribution, rather than optimizing the network structure directly. 
\end{itemize}

Table\ref{tab_ks2} displays the experimental results. Table \ref{tab_ks2} shows the accuracy and network structure's complexity (number of nodes and parameters) of the three algorithms on regression and classification tasks.

\begin{table}[htbp]
\centering
\caption{A comparative analysis is presented, focusing on the accuracy and complexity of the final network structures across three algorithms in both regression and classification tasks.
\label{tab_ks2}}
\resizebox{\textwidth}{64mm}{
\begin{tabular}{cccccccccc}
\toprule[1.45pt]
\toprule[1pt]
\multicolumn{10}{c}{REGRESSION}\\
\toprule
Benchmark& \multicolumn{3}{c}{X-Net}&\multicolumn{3}{c}{Kronecker NN}& \multicolumn{3}{c}{Shinichi\textsuperscript{\cite{shirakawa2018dynamic}}}\\ 
\toprule
&$R^2$& Nod & Para & $R^2$ &Nod & Para  & $R^2$& Nod& Para  \\ 
\cmidrule(lr){2-4}
\cmidrule(lr){5-7}
\cmidrule(lr){8-10}

Nguyen-1 &1.0& 5  & 22  &0.99& 12 & 61&0.99& 12 & 52  \\
Nguyen-2 &1.0& 9 & 38  &0.99& 16& 97 & 0.99&15 & 79\\
Nguyen-3 &1.0& 14 & 58 &0.99& 14 & 78 & 0.99&11& 47\\
Nguyen-4 &0.99& 20 & 82 &0.99& 24 &193 &0.99&18 & 115\\
Nguyen-5 &0.99 & 5 & 16 & 0.99&28 &253 & 0.99&30 & 289\\
Nguyen-6 &0.99&  6 & 18 &0.99& 8 & 33 &0.99& 7 & 27 \\
Nguyen-7 &0.99& 5 & 16 &0.99& 10 & 46 &0.99& 8 &37 \\
Nguyen-8 &1.0& 1 & 4 &0.99& 16 & 97 & 0.99&12& 61\\
Nguyen-9 &1.0& 7 & 24 &0.99& 28 &253 &0.99&21 &139\\
Nguyen-10 &1.0& 4 & 14 &0.99& 80 & 1761 &0.99& 56& 919\\
Nguyen-11 &1.0& 3 & 10 & 0.99& 20 & 141 &0.99&12 & 65\\
Nguyen-12 &0.99& 9 & 40 & 0.99& 12 & 61& 0.99&14& 73\\ 
\cmidrule(lr){1-1}
\cmidrule(lr){2-4}
\cmidrule(lr){5-7}
\cmidrule(lr){8-10}
Average &\textbf{0.996} & \textbf{7.33} & \textbf{28.50} & \textbf{0.990}& \textbf{21.5 }& \textbf{256.17}&\textbf{0.990} &\textbf{ 18} & \textbf{158.58}\\

\toprule
\toprule
\multicolumn{10}{c}{CLASSION}\\
\toprule
Benchmark& \multicolumn{3}{c}{X-Net}&\multicolumn{3}{c}{Kronecker NN}& \multicolumn{3}{c}{Shinichi\textsuperscript{\cite{shirakawa2018dynamic}}}\\ 
\toprule
&Acc& Nod & Para & Acc &Nod & Para  & Acc& Nod& Para  \\
\cmidrule(lr){2-4}
\cmidrule(lr){5-7}
\cmidrule(lr){8-10}
Iris &98.7\%& 28 & 112 &98.3\%& 66 & 798 &98.6\%& 68& 823\\
Mnist(6-dim) &89.4\%& 65 & 244  &88.8\%& 276 & 13328 &88.9\%&256 & 12194\\
Mnist &99.5\%& 816 & 3084 &99.2\%& 682 & 228073& 99.0\%&658& 201291\\ 
Fashion-MNIST(6-dim)&76.2\%& 122 & 486 & 75.3\%& 364 & 22432 & 75.2\%& 344 & 21276\\
Fashion-MNIST&94.1\%&  1066& 3884 & 94.5\%& 1167 & 505159 & 94.3\%& 1088 & 448425\\
CIFAR-10(6-dim) &26.4\%& 206 & 764 & 22.7\%& 398 & 24338 & 25.2\%& 344 & 19784\\
CIFAR-10 &46.8\%& 2733 & 10072 & 44.8\%& 1932 & 743355 & 45.1\%& 1894 & 726110\\ 
\cmidrule(lr){1-1}
\cmidrule(lr){2-4}
\cmidrule(lr){5-7}
\cmidrule(lr){8-10}
Average &\textbf{75.9\%} & \textbf{719.43}& \textbf{2663.71 } &\textbf{74.8\%}&\textbf{697.86}&\textbf{219640.43} &\textbf{75.2\%}&\textbf{ 497} & \textbf{204271.85}\\
\toprule
\end{tabular}
}
\end{table}

\section{X-Net Powers Scientific Discovery}  
\begin{figure*}[htbp] 
    \centering
	  \subfloat[Performance of the Ada-$\alpha$ Activation Function]{
        \includegraphics[width=0.3\linewidth]{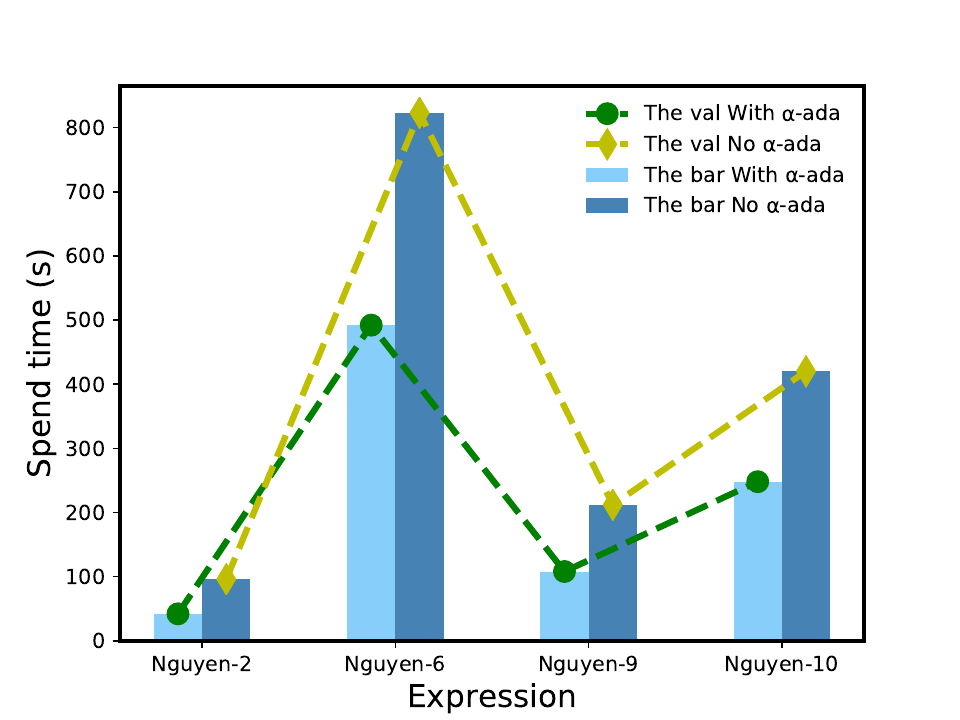}
    \label{1b}}
      \subfloat[Boston Housing Price Data Correlation Matrix.]{
       \includegraphics[width=0.3\linewidth]{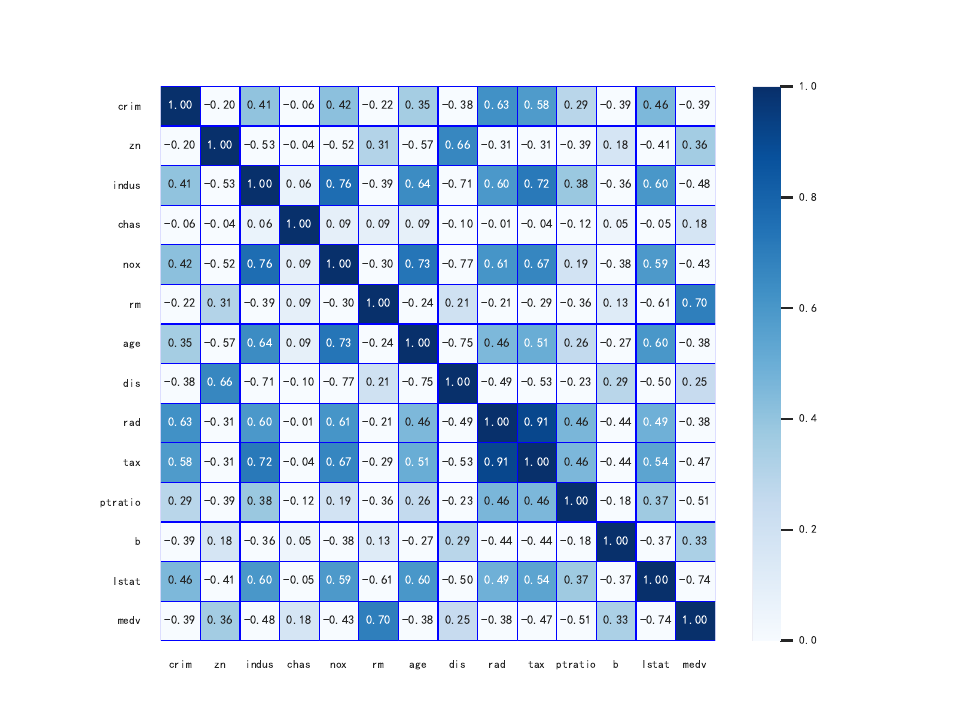}
    \label{1c}}\\
    \subfloat[Predicting housing prices using the 'RM']{
        \includegraphics[width=0.3\linewidth]{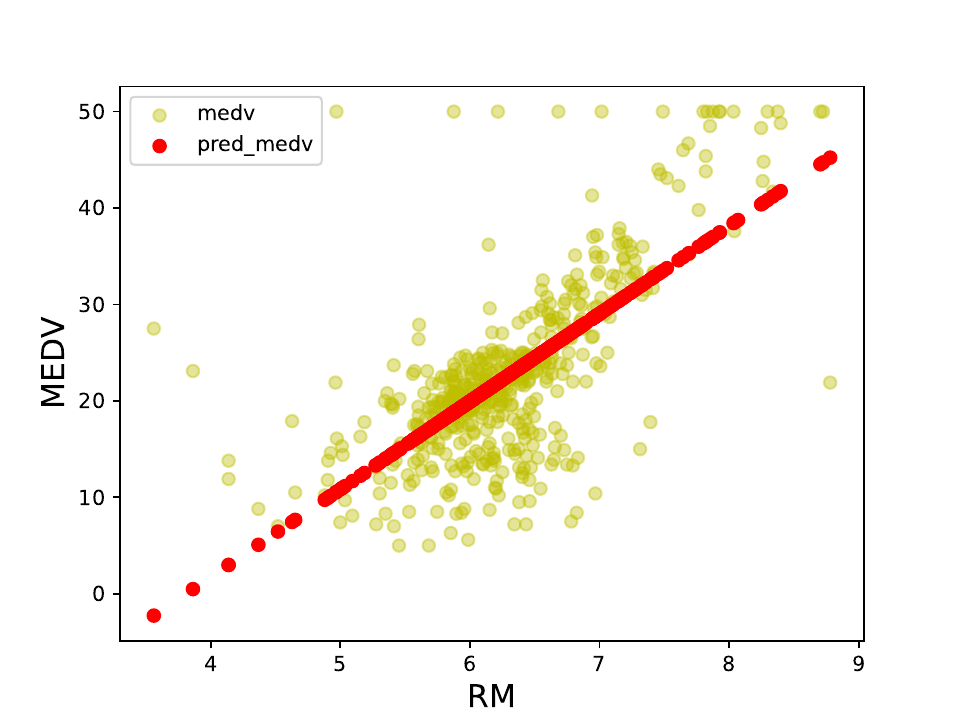}
    \label{1d}}
    \subfloat[Predicting housing prices using the 'LSTAT']{
        \includegraphics[width=0.3\linewidth]{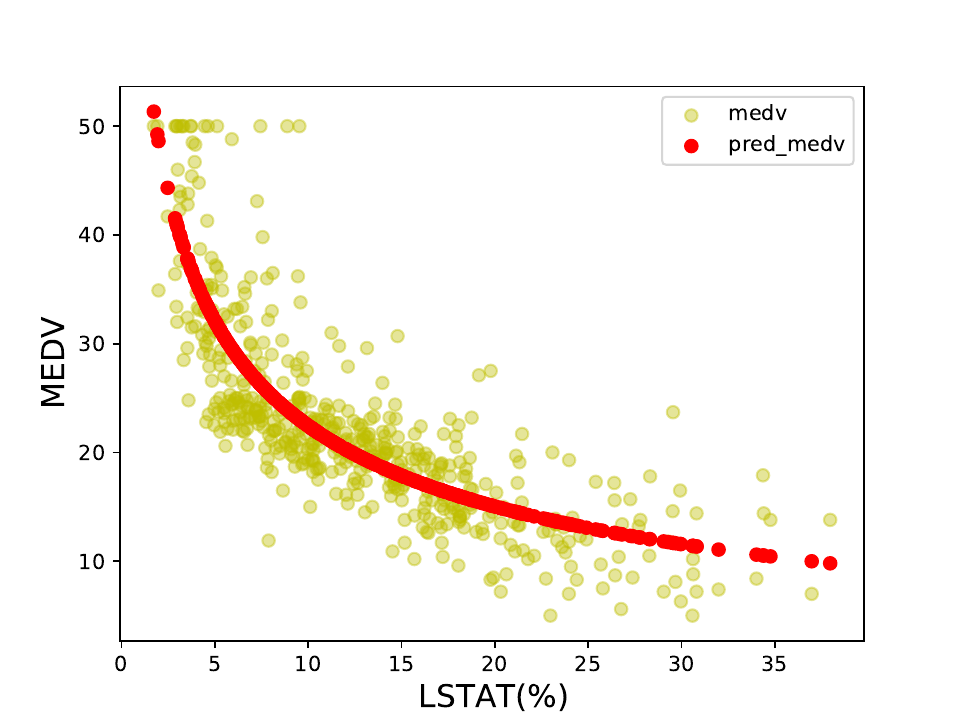}
    \label{1e}}
	 \subfloat[Predicting housing prices using 'RM' and 'LSTAT']{
        \includegraphics[width=0.3\linewidth]{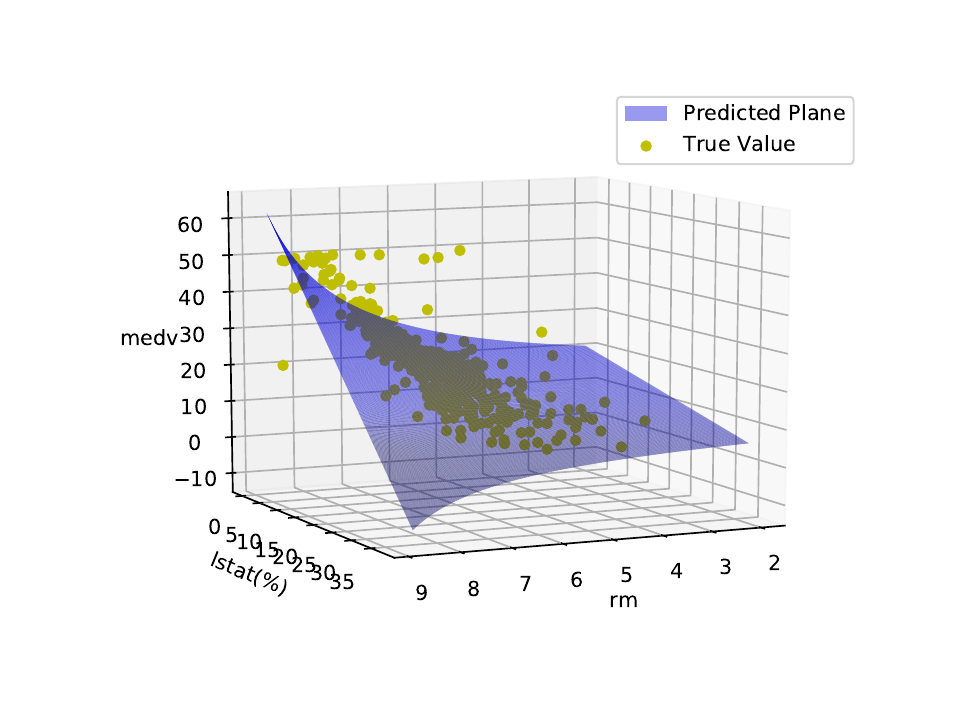}
    \label{1f}}
    \\
	  \caption{Figure a depicts the change in algorithmic efficiency before and after the use of ada-$\alpha$; Figure b presents the correlation matrix among different variables in the Boston housing price prediction; Figure c demonstrates the schematic representation of housing price predictions using the variable `RM'. It can be deduced that the housing prices are directly proportional to the variable `RM', which is consistent with the findings presented in Figure c; Figure d displays the results of predicting housing prices using the variable `LSTAT'. Figure e showcases the schematic representation of predicting housing prices using both `RM' and `LSTAT'.}
	  \label{fig2} 
\end{figure*}

\subsection{Economic Modeling}
The Boston house price forecast data set \textsuperscript{\cite{boston}} is a classic data set in machine learning. Although it is simple, it can test the comprehensive performance of the algorithm well.
Each sample in the Boston housing price data set is composed of 12 characteristic variables such as 'CRIM', 'ZN', and housing price MEDV.
The thermal map of the correlation coefficient between each variable is shown in the figure\ref{1c}.
From the heat map \ref{1c}, we can find that the variable 'RM' (the number of rooms per house) has the highest correlation coefficient with the MEDV(housing price), which is 0.70, showing a positive correlation with the housing price. In contrast, the variable 'LSTAT' (how many people in the area are classified as low-income) has a minimum correlation coefficient of -0.74 
with housing prices, showing a negative correlation with housing prices.

We use RM to forecast the house price MEDV, and the final formula was as follows\eqref{rm}: 
\begin{equation}
\label{rm}
\displaystyle MEDV = 9.10 RM - 34.67 (RM\geq2)
\end{equation}

The predicted results are shown in Fig. \ref{1d}. From Formula \ref{rm}, we can find that the final expression obtained by our algorithm can well reflect the positive correlation between RM and housing price. \\
We used LSTAT as input to predict housing prices using X-Net and finally got formula \eqref{lstat}, from which we could see that the formula also well reflected the negative correlation between LSTAT and MEDV. Show in Fig \ref{1e}.
\begin{equation}
\label{lstat}
\displaystyle MEDV = \frac{81.39}{(LSTAT + 0.44)^{0.44}} - 6.54 (0\leq LSTAT \leq 100)
\end{equation}
Finally, we use RM and LSTAT as inputs to predict housing prices, and we get formula \eqref{medv4}, the predicted results are shown in Fig. \ref{1f}
\begin{equation}
\label{medv4}
\displaystyle MEDV = \frac{15.6 (0.48  RM - 13.21)}{2.36 LSTAT - 24.53} + 9.97
\end{equation}
\begin{figure*}[htbp] 
    \centering
    \subfloat[Solar energy power generation fitting result]{
        \includegraphics[width=0.3\linewidth]{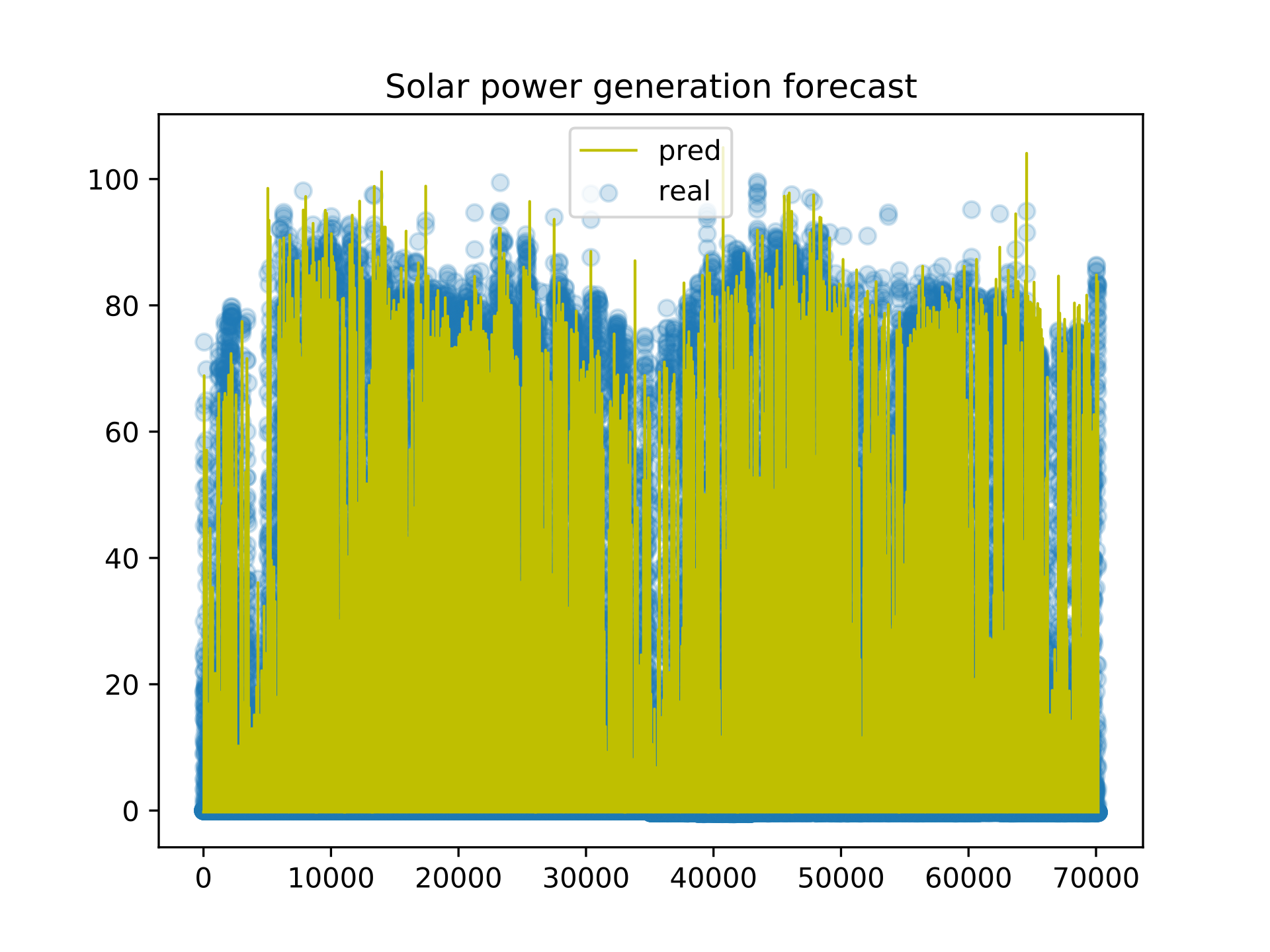}
    \label{2g}}
    \subfloat[Figure g after sorting by $y_{pre}$]{
       \includegraphics[width=0.3\linewidth]{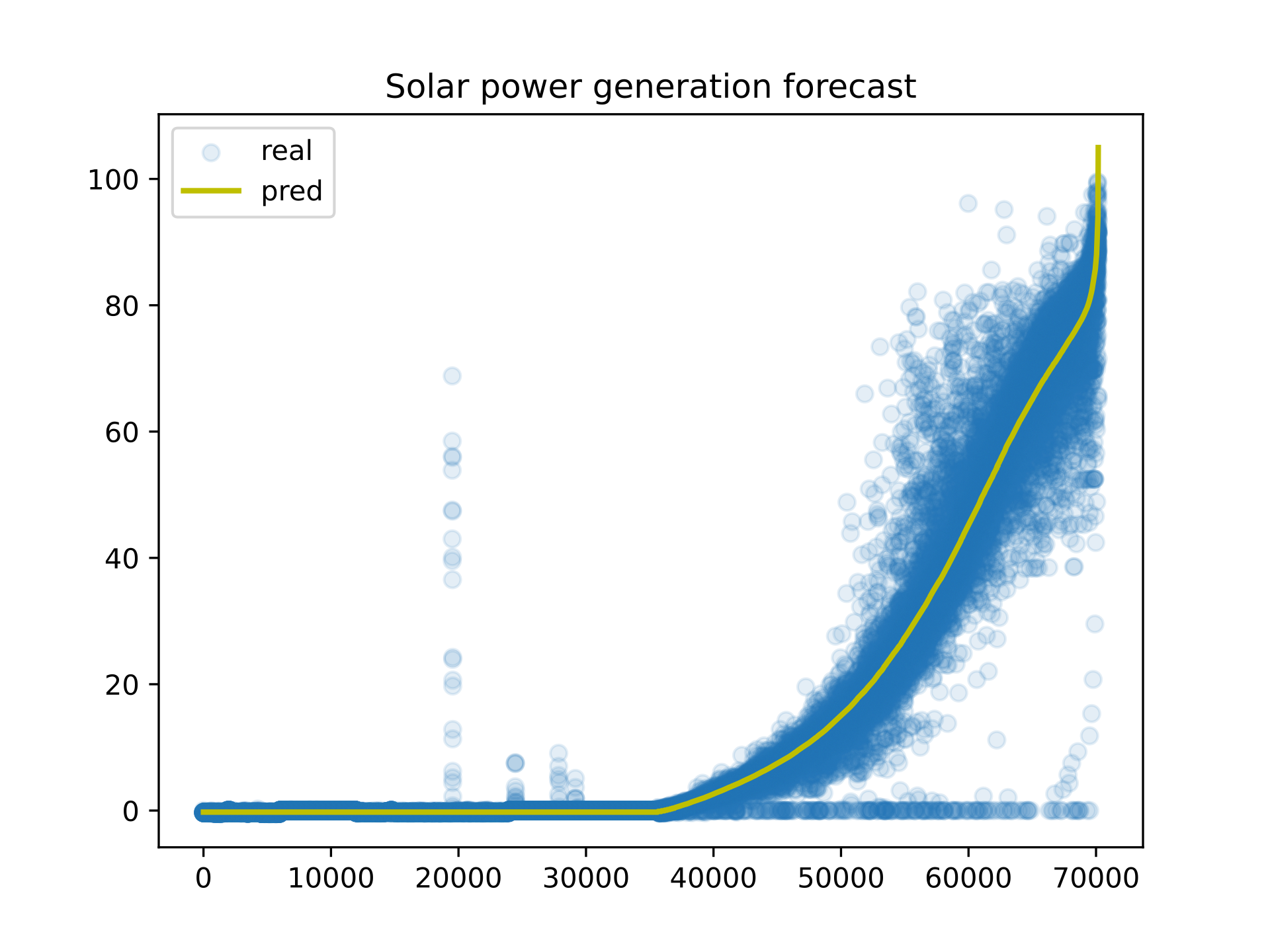}
    \label{2h}}
      \\
      \subfloat[Wind energy power generation fitting result]{
        \includegraphics[width=0.3\linewidth]{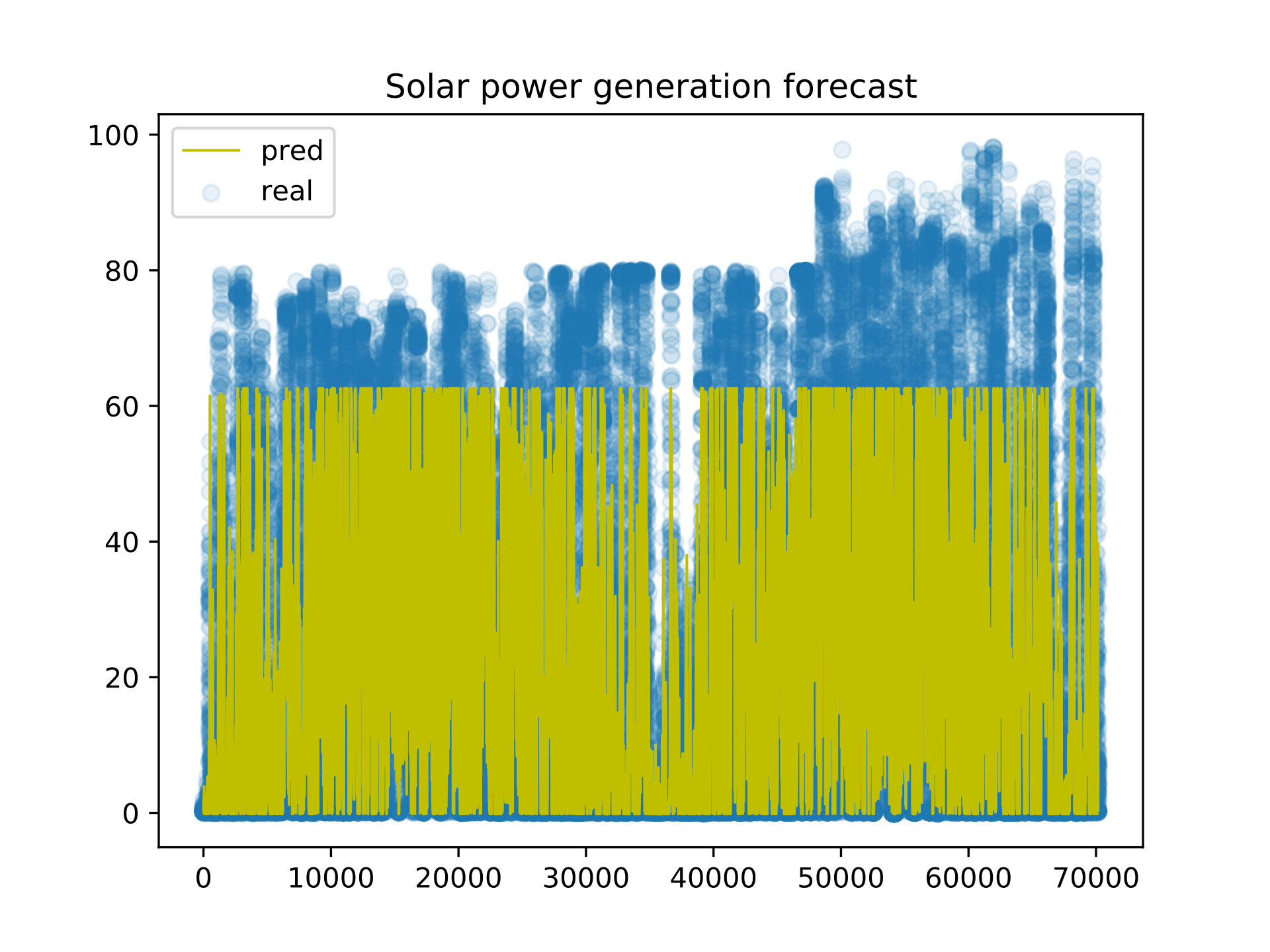}
    \label{2i}}
      \subfloat[Figure i after sorting by $y_{pre}$]{
        \includegraphics[width=0.3\linewidth]{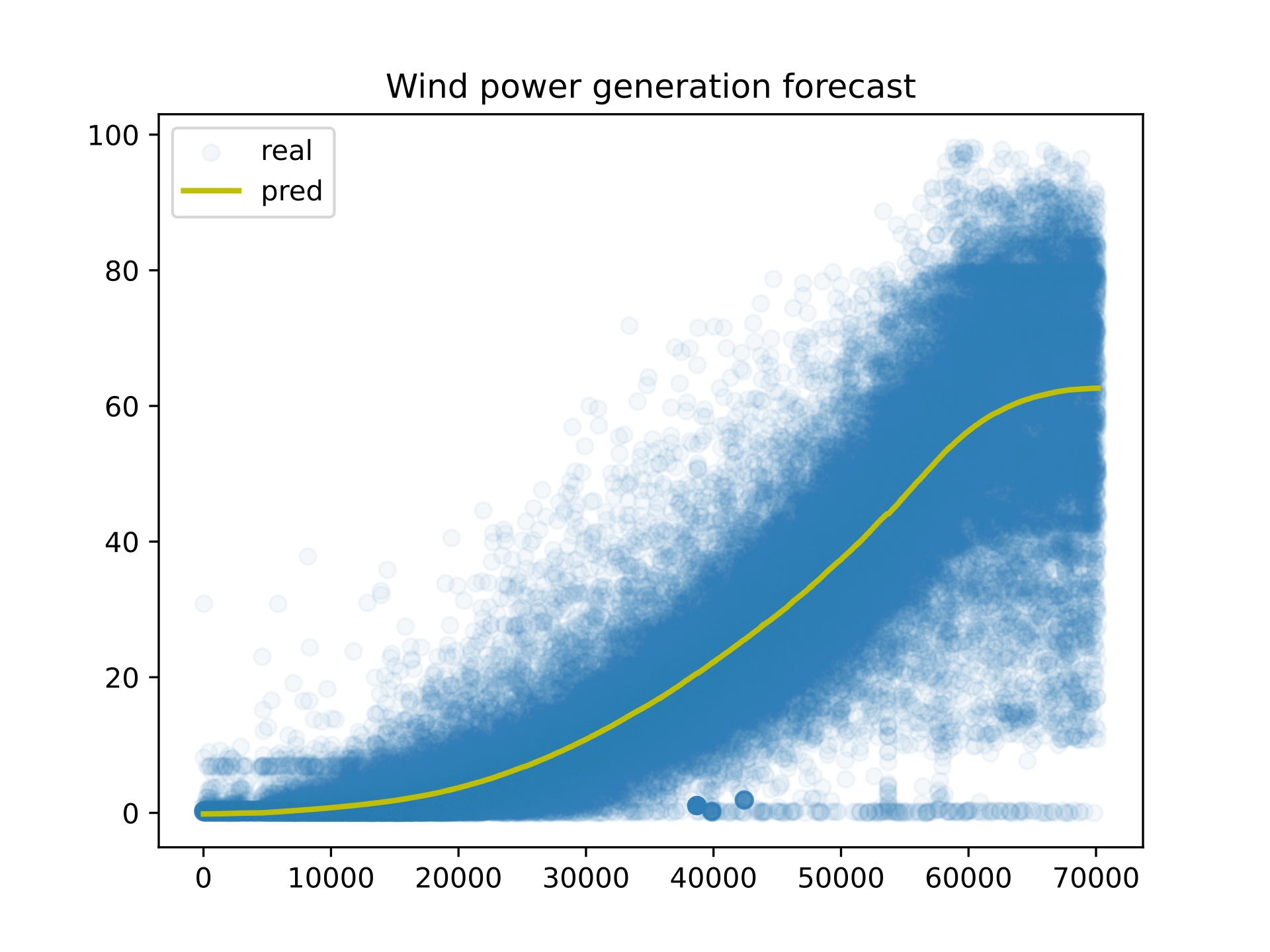}
    \label{2j}}
	  \subfloat[The correlation matrix in wind energy prediction ]{
        \includegraphics[width=0.3\linewidth]{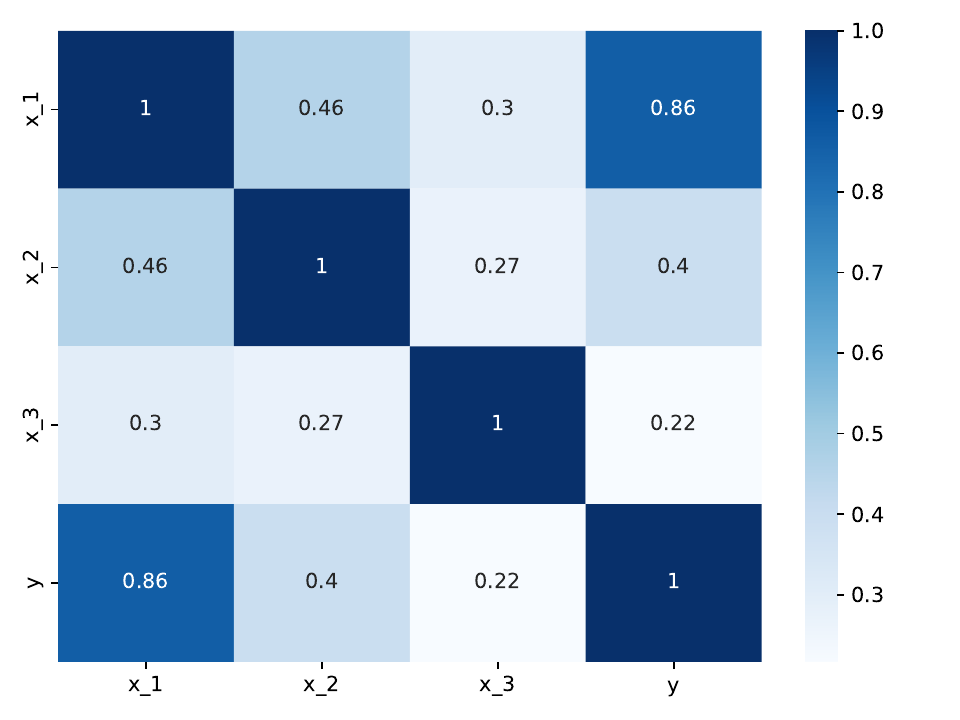}
    \label{2k}}\\
	  \caption{Figure a and Figure b show the prediction results of Formula \ref{equ8} for solar power generation; Figure c and Figure d display the fitting results of Formula \ref{equ9} for wind power generation data; Figure e represents the correlation coefficient matrix of the variables used in the wind power generation data.}
	  \label{fig2} 
\end{figure*}
From Formula \ref{medv4}, we can clearly find that the housing price is directly proportional to the variable RM and inversely proportional to the variable LSTAT. That is the more rooms in the house, the more expensive the house, the more low-income people in the community, and the lower the price. The result of this prediction is completely in line with the facts. The above results prove that our algorithm can well reflect the relationship between each variable and the result even for multiple variables. 

\subsection{Modeling in energy science
(Solar And Wind Power Generation Forecasting)\textsuperscript{\cite{sunwind}}}
Accurate solar and wind power generation predictions are crucial for advanced electricity scheduling in energy systems. We used the solar and wind energy data provided by the State Grid Corporation of China to model the data. This data set consists of data directly collected from renewable energy stations, including generation and weather-related data from six wind farms and eight solar power stations distributed across different regions of China over the course of two years (2019-2020), collected every 15 minutes. The solar power generation data contains three main variables: Total solar irradiance($x_1$), Air temperature($x_2$), and Relative humidity($x_3$). The solar energy generation prediction model obtained through training X-Net can be represented by equation \ref{equ8}, as follows:
\begin{equation}
\label{equ8}
\displaystyle Y = x_1 (0.00044  x_3(-0.00003x_1x_2 + 0.31) + 0.093) - 0.2466
\end{equation}
The fitted curves of the sun power are shown in Figure \ref{2g} and \ref{2h}. Note: Figure \ref{2h} underwent the same data sorting process as Figure \ref{2b}.\\
The wind power generation data also mainly includes three variables, Wind speed at the height of the wheel hub($x_1$), Wind direction at the height of the wheel hub($x_2$), and Air temperature($x_3$). The formula model obtained by X-Net for wind power generation using the above three variables as input can be represented by equation \ref{equ9}, as follows:
\begin{equation}
\label{equ9}
\displaystyle Y = x_1(0.0123x_1 + \frac{0.0123(x_1-1.53)}{0.00049  x_1(x_1 - 15.93)+0.049)})
\end{equation}
The fitted curves of the wind power are shown in Figure \ref{2i} and \ref{2j}. Note: Figure \ref{2j} underwent the same data sorting process as Figure \ref{2b}.
From Equation \ref{equ9}, we can see that X-Net has modeled the data well using only the first variable (wind speed) among the three variables. This is consistent with our understanding that wind power generation is mainly related to wind speed, which also aligns with reality. Furthermore, from the correlation coefficient heatmap \ref{2k}, we can clearly see that the correlation between $x_1$ (wind speed) and $y$ (power generation) is as high as 0.86.

\section{Ada-$\alpha$ function performance analysis}
In the process of neural network training, whether the step size is appropriate directly affects the speed of network convergence. If the step size is excessively large, the optimal solution may be skipped, If the step size is too small, the convergence rate may be extremely sluggish, and the optimal solution may not be obtained within a predetermined number of iterations. Therefore, to improve network efficiency, we design a step-size dynamic adaptive function Ada-$\alpha$. In order to test its performance, we selected four functions 'Nguyen-2', 'Nguyen-6', 'Nguyen-9', and 'Nguyen-10' from the Nguyen data set to test the Ada-$\alpha$. Using the Ada-$\alpha$ function, we execute each of the formulas ten times, record the amount of time required to fully recover the formula, and then calculate an average. Then do the same for the case without the Ada-$\alpha$ function. The specific bar graph is depicted in Fig. \ref{1b}. We can observe that when the Ada-$alpha$ function is implemented, the network's convergence speed will increase dramatically.


\end{document}